\documentclass{article}

\usepackage{microtype}
\usepackage{graphicx}
\usepackage{subfigure}
\usepackage{booktabs} 
\usepackage{enumitem}
\usepackage{multirow}
\usepackage{adjustbox}
\usepackage{tabularx}

\usepackage{colortbl}
\usepackage{xcolor}
\usepackage{array}
\usepackage{pifont}
\usepackage{hyperref}
\usepackage{makecell}

\usepackage[accepted]{icml2025}

\usepackage{amsmath}
\usepackage{amssymb}
\usepackage{mathtools}
\usepackage{amsthm}

\usepackage[capitalize,noabbrev]{cleveref}

\theoremstyle{plain}

\theoremstyle{definition}

\theoremstyle{remark}

\usepackage[textsize=tiny]{todonotes}

\icmltitlerunning{MM-OpenFGL: A Comprehensive Benchmark for Multimodal Federated Graph Learning}

\begin{document}

\twocolumn[
\icmltitle{MM-OpenFGL: A Comprehensive Benchmark for\\ Multimodal Federated Graph Learning}



\icmlsetsymbol{equal}{*}

\begin{icmlauthorlist}
\icmlauthor{Xunkai Li}{bit}
\icmlauthor{Yuming Ai}{bit}
\icmlauthor{Yinlin Zhu}{sysu}
\icmlauthor{Haodong Lu}{bit}
\icmlauthor{Yi Zhang}{sdu}
\icmlauthor{Guohao Fu}{bit}
\icmlauthor{Bowen Fan}{bit}
\icmlauthor{Qiangqiang Dai}{bit}
\icmlauthor{Rong-Hua Li}{bit}
\icmlauthor{Guoren Wang}{bit}
\end{icmlauthorlist}

\icmlaffiliation{bit}{Beijing Institute of Technology, Beijing, China}
\icmlaffiliation{sysu}{Sun Yat-sen University, Guangzhou, China}
\icmlaffiliation{sdu}{Shandong University, Weihai, China}

\icmlcorrespondingauthor{Rong-Hua Li}{lironghuabit@126.com}

\icmlkeywords{Machine Learning, ICML}

\vskip 0.3in
]



\printAffiliationsAndNotice{}  

\begin{abstract}

Multimodal-attributed graphs (MMAGs) provide a unified framework for modeling complex relational data by integrating heterogeneous modalities with graph structures. While centralized learning has shown promising performance, MMAGs in real-world applications are often distributed across isolated platforms and cannot be shared due to privacy concerns or commercial constraints. Federated graph learning (FGL) offers a natural solution for collaborative training under such settings; however, existing studies largely focus on single-modality graphs and do not adequately address the challenges unique to multimodal federated graph learning (MMFGL). To bridge this gap, we present MM-OpenFGL, the first comprehensive benchmark that systematically formalizes the MMFGL paradigm and enables rigorous evaluation. MM-OpenFGL comprises 19 multimodal datasets spanning 7 application domains, 8 simulation strategies capturing modality and topology variations, 6 downstream tasks, and 57 state-of-the-art methods implemented through a modular API. Extensive experiments investigate MMFGL from the perspectives of necessity, effectiveness, robustness, and efficiency, offering valuable insights for future research on MMFGL.

\end{abstract}

\section{Introduction}
\label{sec: introduction}

Multimodal-attributed graph (MMAG) serves as a powerful modeling technique for relational data in which nodes are associated with heterogeneous modalities (e.g., text and images) and edges encode interactions among entities. By jointly modeling complementary multimodal signals, recent MMAG methods have achieved strong performance across a wide range of applications. However, these advances predominantly rely on a centralized learning paradigm, implicitly assuming that all modalities and graph structures are accessible to a single graph learning system.

However, such an assumption is often violated in practice, where MMAGs are naturally distributed across multiple data owners due to privacy constraints or commercial competition. For example, different platforms independently maintain social graphs over disjoint user populations, where posts combine textual content with visual attachments, yet cross-platform data sharing is prohibited. As a result, existing centralized MMAG frameworks are fundamentally limited in their ability to exploit decentralized multimodal knowledge, hindering global cross-modal alignment and the discovery of collective multimodal intelligence.

\vspace{0.1cm}

To address these challenges, a natural recourse is the adoption of federated graph learning (FGL), a paradigm that has demonstrated significant efficacy across diverse decentralized graph learning scenarios. By facilitating the collaborative training of graph models through the exchange of model updates instead of raw data, FGL preserves data privacy while simultaneously capitalizing on decentralized knowledge. Nevertheless, existing FGL frameworks are predominantly tailored for single-modality attributed graphs characterized by homogeneous node features. In such settings, the primary learning obstacles arise from structural heterogeneity or shifts in label distributions. Extending FGL to multimodal environments gives rise to multimodal federated graph learning (MMFGL), a transition that is fundamentally non-trivial. As evidenced by our empirical study in Sec.~\ref{sec: exp_necessity} (\textbf{Q2}), naive adaptations of conventional FL or FGL methodologies fail to reconcile cross-modal semantic conflicts and cross-client structural mismatches, occasionally yielding results inferior to those of isolated training. Consequently, there is an urgent necessity to establish a systematic foundation for MMFGL.

\begin{table*}[t]
\centering
\caption{Comparison between existing FGL benchmarks and our proposed MM-OpenFGL for MMFGL.}

\setlength{\tabcolsep}{6pt}
\renewcommand{\arraystretch}{1.7}
\footnotesize

\begin{tabular*}{\textwidth}{
@{\extracolsep{\fill}}
c c c c c c
@{}
}
\toprule
\textbf{Benchmark} &
\textbf{Multimodal} &
\textbf{Simulation} &
\textbf{Algorithms} &
\textbf{Downstream Tasks} &
\textbf{Evaluations} \\
\midrule
Bkd-FedGNN~\cite{liu2025bkd} & \ding{55} & 2 (Label) & 4 & 2 Graph-Level & 6 \\
OpenFGL~\cite{Li2024OpenFGLAC} & \ding{55} & 4 (Topo. \& Label) & 18 & 5 Graph-Level & 11 \\
FedGraph~\cite{yao2024fedgraph} & \ding{55} & 2 (Label) & 14 & 3 Graph-Level & 7 \\
\midrule
MM-OpenFGL &
\checkmark &
8 (2 Modal. \& Topo. \& Label) &
57 &
\makecell[c]{4 Graph-Level \\ 5 Modality-Level} &
14 \\
\bottomrule
\end{tabular*}
\label{benchmark_comp}
\end{table*}

\vspace{0.1cm}
Despite its importance, MMFGL still remains largely underexplored, primarily due to three fundamental gaps: \ding{182} \underline{\textit{Lack of Problem Formalization:}} There is no unified formulation that systematically characterizes different MMFGL settings and tasks in the current literature. This lack of clarity prevents researchers from pinpointing the unique challenges arising from the intersection of MMAGs and decentralized learning. \ding{183} \underline{\textit{Lack of Benchmarking System:}} Current FGL libraries (Table~\ref{benchmark_comp}) lack support for multimodal graph datasets, modality-specific simulation processes, and related learning tasks. A standardized, accessible benchmark featuring open datasets and reference implementations is essential to foster the development of MMFGL methodologies. \ding{184} \underline{\textit{Lack of Directional Insights:}} Systematic empirical studies on MMFGL are largely absent. Without experimental evidence regarding how factors like modality heterogeneity or cross-modal alignment influence federated training dynamics, it remains unclear which challenges are most critical. This empirical void hinders the principled design and rigorous evaluation of MMFGL algorithms.

To this end, we propose MM-OpenFGL, the first comprehensive benchmark specifically designed to formalize the MMFGL paradigm. MM-OpenFGL can be viewed as a systematic extension of OpenFGL: while OpenFGL focuses on single-modal graphs and provides a foundation for FGL evaluation, MM-OpenFGL inherits its modular API but substantially extends its scope to MMAGs and federated graph foundation model (Appendix~\ref{appendix: Comparison with Related Benchmarks} for details). Specifically, MM-OpenFGL integrates 19 multimodal graph datasets across 7 application domains, 8 simulation strategies, 57 state-of-the-art algorithms, and 9 graph- and modality-based downstream tasks providing a unified and scalable framework for benchmarking MMFGL methods. Using this benchmark, we conduct an extensive empirical study of existing baselines and derive 10 key insights, covering Necessity, Effectiveness, Robustness, and Efficiency.

\noindent \textbf{Our Contributions.} \ding{182} \textbf{Problem Formalization.} To the best of our knowledge, we are the first to introduce the multimodal federated graph learning (MMFGL) problem. \ding{183} \textbf{Comprehensive Benchmark.} We present MM-OpenFGL, a pioneering MMFGL benchmark spanning 19 multimodal datasets across 7 domains. It defines 8 distributed simulation settings based on modality, label and topology, incorporates 57 algorithms, and evaluates them from four perspectives: necessity, effectiveness, robustness, and efficiency. \ding{184} \textbf{Valuable Insights.} Through extensive empirical studies, we derive 10 conclusions, outlining promising research directions for further MM-FGL community. \ding{185} \textbf{Community Resources.} We release an open-source library and a curated repository including a systematic literature review to lower the entry barrier for MMFGL. All code, datasets, and tutorials are available at \textcolor{red}{\url{https://anonymous.4open.science/r/TEST-SA7D7A}}.

\begin{table*}[t]
\caption{An overview of our proposed MM-OpenFGL benchmark.}
\label{tab: An overview of OpenFGL}
\resizebox{\textwidth}{!}{
\begin{tabular}{c|c}

\toprule

\textcolor{blue}{\textbf{Scenarios}} & \multicolumn{1}{c}{\cellcolor{blue!15}{{\textit{\textbf{Data}}}}} \\ 
     \midrule[0.3pt]
     
Datasets     & Movies, Cloth, Ele-fasion, Bili cartoon, MultiMET facebook, PixelRec50K...    \\
Encoder & Qwen2.5-3B-Instruct, Qwen2-7B-Instruct, Clip-Vit-Large, Llama-3.2-11B-Vision-Instruct,... \\

Simulation   & Modality-IID / Modality-NonIID, Topology-Available / Topology-Unavailable, Label-IID / Label-NonIID               \\
Tasks        & Graph-Level(Node Classification, Link Prediction,...), Modality-Level(Modality Match, Modality Generation,...)       \\ 

\midrule[0.3pt]

\textcolor{red}{\textbf{Method}}                     & \multicolumn{1}{c}{\cellcolor{red!15}{{\textit{\textbf{Algorithms}}}}}                                                                                                                     \\ \midrule[0.3pt]
GNN                  & \multicolumn{1}{c}{GCN, GAT, GraphSAGE, SGC, GIN, ChebNet, MMA, RevGAT, BUDDY, ...}                                                                                           \\
MM-GNN & \multicolumn{1}{c}{MMGCN, MGAT, GSMN, MGNet, GraphMAE2, MMGCL, LGMRec, COGMEN, DMGC, ...}\\

Standard-FL                   & \multicolumn{1}{c}{FedGM, FedSPA, FedIIH, FedSSP, S2FGL, FedLap, FCGL, FGC, FedGLS, FedDEP, FedAGHN, ... }                                                                       \\
Heterogeneous-FL                  & \multicolumn{1}{c}{FedTGP, FML, FIARSE, ReeFL, FedTSA, MH-pFLID, PEPSY, FedMVP, FedMAC, FedMM, ... }                                                           \\ 
GFM & \multicolumn{1}{c}{GFT, OFA, UniGraph, GQT, GraphCLIP, AnyGraph, GFSE, SwapGT, RAGraph} \\

\midrule[0.3pt]
\textcolor[rgb]{0.0, 0.5, 0.0}{\textbf{Experiment}}                      & \multicolumn{1}{c}{\cellcolor{green!15}{{\textit{\textbf{Evaluations}}}}}                                                                                                                    \\ \midrule[0.3pt]
Data Analysis    & \multicolumn{1}{c}{Feature KL Divergence, Label Distribution, Topology Statistics}      \\
Effectiveness        & \multicolumn{1}{c}{ Accuracy, Precision, F1, AUC-ROC, Recall@K, MRR, CLIP Score, BLEU}                                                 \\
Robustness           & \multicolumn{1}{c}{Noise, Sparsity, Homophily, Federated Scenario Generalization, Privacy Preserve} \\
Efficiency           & \multicolumn{1}{c}{Convergence, Scalability, Communication, FLOPS, Time\&Space Complexity} \\

\bottomrule
\end{tabular}
}
\vspace{0.1cm}
\end{table*}

\section{Problem Statement}
\label{sec: problem statement}
To provide a comprehensive evaluation of MMAGs in federated scenarios, we classify training pipelines into two paradigms: the End-to-End (standard FGL) pipeline for task-specific collaboration and the Two-Stage (graph foundation model) pipeline for task-agnostic representation learning, which are detailed as follows:

\subsection{End-to-End Pipeline}

This pipeline follows the classical federated learning approach, training a task-specific model from scratch through iterative communication. Following the OpenFGL protocol, the training process at communication round $t$ involves four steps:
\ding{182} \textbf{Receive Message}: The server distributes the current global model to selected clients. Each client initializes its local model with the received parameters: $\mathbf{w}^T_k \leftarrow \mathbf{\hat w}^T$;
\ding{183} \textbf{Local Update}: Each client $k$ optimizes the model on its private multimodal graph using task-specific supervision, targeting the local objective;
\ding{184} \textbf{Upload}: Clients send their updated parameters or gradients back to the server;
\ding{185} \textbf{Aggregate}: The server aggregates the updates to form the new global model, e.g., via FedAvg: $\mathbf{\hat w}^{T+1} \leftarrow \frac{1}{D}\sum^K_{k=1} D_k\mathbf{w}^{T+1}_k$, where $D$ is the total number of samples.

\subsection{Two-Stage Pipeline}

This pipeline follows a pre-train-then-fine-tune strategy to leverage graph foundation models. It evaluates both the generalization of the pre-trained backbone and its transferability to local tasks.
\ding{182} \textbf{Federated Pre-training (Stage 1).} Clients collaboratively train a global graph encoder using self-supervised tasks such as link prediction, masked feature reconstruction, or cross-modal contrastive learning. The communication cycle mirrors the four-step process above, but the goal is to learn generic structural and semantic representations;
\ding{183} \textbf{Local Fine-tuning (Stage 2).} After pre-training, the global model acts as a foundation backbone. Each client downloads the model and fine-tunes it on local tasks, allowing adaptation to the client-specific graph distribution without requiring further server communication.

\section{Benchmark Design}
\label{sec: benchmark design}

\subsection{Decentralized MMAGs Simulation}
In this section, we describe the construction of the MM-OpenFGL data ecosystem, including the underlying source datasets, the adopted multimodal encoders, and the proposed tri-dimensional federated simulation scenarios.

\vspace{0.05cm}
\textbf{MMAG Datasets.} To ensure robust generalization, MM-OpenFGL curates a comprehensive collection of real-world datasets spanning e-commerce, recommendation systems, social media, and medical imaging. Distinct from existing benchmarks that rely on pre-processed features, we uniquely provide raw multimodal data (text and images) aligned with graph structures. The collection is organized into three complementary families: sequence recommendation graphs (e.g., Bili series), multimodal product graphs (e.g., PixelRec50K), and domain-specific graphs (e.g., social media).

\vspace{0.05cm}
\textbf{Modality Encoder.} To connect unstructured modalities with graph-structured data, we use pre-trained multimodal encoders such as Qwen2-7B-Instruct for node feature extraction. In addition, we have assembled a collection of single-modality encoders. For text, this includes models like Llama-3.2-1B-Instruct and T5. For vision, it includes ViG and DINOv2. Incorporating a range of backbones, from the lightweight Llama-3.2-1B to the high-capacity DINOv2, allows researchers to explore the trade-off between feature quality and computational efficiency in federated scenarios.

\vspace{0.05cm}
\textbf{Simulation Strategies.}
To simulate real-world scenarios, we propose a tri-dimensional simulation strategy that yields $2\times2\times2$ federated scenarios, including the modality, topology, and label perspectives. We organize and introduce these perspectives by heterogeneity dimension as follows:

\ding{182} \textbf{Modality-IID and Modality-NonIID.} To tackle data fragmentation in multimodal environments, we propose the modality-NonIID strategy. This approach uses a Dirichlet distribution to simulate missing modalities, capturing real-world factors such as privacy restrictions or diverse user behavior. By partitioning modalities across clients, the model is encouraged to learn robust representations from disjoint or partial modality views.

\ding{183} \textbf{Topology-Available and Topology-Unavailable.}
This dimension addresses structural data constraints and privacy, such as hidden edges in financial networks. We introduce two contrasting strategies: topology-unavailable, which removes original edges and artificially reconstructs the topology using methods like SBM~\cite{sbm} or RDPG~\cite{rdpg}, and topology-available, which preserves the inherent graph structure. Notably, the topology-unavailable strategy enhances scalability by effectively modeling unstructured multimodal data (e.g., independent collections of images and texts) that naturally lack explicit topological connections.

\ding{184} \textbf{Label-IID and Label-NonIID}
This dimension captures label-induced statistical heterogeneity. We model-NonIID settings using standard partitioning methods (e.g., Louvain~\cite{Blondel_louvain} and Metis~\cite{karypis1998metis}), while the IID setting enforces label-balanced partitions across clients.

Integrating modality heterogeneity, topological availability, and label properties into eight strategies, MM-OpenFGL establishes the first unified framework for multimodal federated graph learning. This design enables stress testing under realistic complexities, revealing vulnerabilities and guiding robust, privacy-preserving, theoretically sound solutions.

\textbf{Downstream Tasks.} Unlike traditional graph benchmarks that focus solely on structural reasoning, MM-OpenFGL introduces a dual-level task evaluation system designed to assess both topological utilization and cross-modality semantic understanding. Regarding graph-level tasks, MM-OpenFGL includes the fundamental objectives of node classification and link prediction. Also, MM-OpenFGL incorporates modality-level tasks to specifically target the semantic interplay between modalities, which is often neglected in standard FGL evaluations. This category encompasses modality matching, modal retrieval, modal alignment, and modality generation.

\subsection{Algorithm Taxonomy}
To provide a holistic evaluation landscape, we establish a comprehensive algorithm taxonomy that systematically categorizes the diverse methodologies integrated into MM-OpenFGL. This taxonomy spans the spectrum from end-to-end traditional learning paradigms to 2-stage pipeline.

\textbf{MM-GNN.}
This category is a network architecture specifically designed for multimodal-attributed graph learning, and also serves as the architectural backbone for local client training. It encompasses state-of-the-art networks specifically designed to fuse graph topology with multimodal features within a centralized setting. By integrating representative models such as MM-GCN~\cite{mmgcn} and MGAT~\cite{mgat}, we establish critical performance upper bounds and robust local baselines, allowing researchers to isolate the impact of federated constraints on multimodal fusion efficacy.

\textbf{Standard FL.}
Standard FL approaches are designed for isomorphism scenarios where all clients share an identical architecture. This pillar is further divided into general FL methods and specialized federated graph learning algorithms. The former includes classic optimization frameworks like FedAvg, FedProx~\cite{fedprox}, and SCAFFOLD~\cite{karimireddy2020scaffold}, originally developed for vision or general tasks but adapted here for graph contexts. The latter comprises algorithms explicitly tailored to handle structural issues, such as FedSPA~\cite{Tan2025FedSPAG} and FedIIH~\cite{fediih}, representing the current standard for distributed graph optimization.

\textbf{Heterogeneous FL.}
Heterogeneous federated learning approaches are designed to address the challenge of system heterogeneity. Recognizing that real-world clients often possess varying computational resources requiring distinct model architectures and novel data heterogeneity problem encountered in modality-NonIID scenarios, we include methods that support model heterogeneity. This category spans both heterogeneous FGL methods like MH-pFLID~\cite{mh-pflid}, and heterogeneous FL methods adapted from the computer vision domain like PEPSY~\cite{pepsy} and FedMVP~\cite{fedmvp}. This inclusion is vital for benchmarking adaptability in resource-constrained and modality-NonIID scenarios.

\textbf{Graph Foundation Model.}
To reflect the frontier of the field, MM-OpenFGL integrates graph foundation models. Moving beyond traditional supervision, this category incorporates large-scale pre-trained models tailored for graph tasks, such as GFT~\cite{gft} and OFA~\cite{ofa}. By evaluating these foundation models, we aim to investigate the potential of Graph-LLM paradigms in a multimodal federated context, specifically testing their capabilities on complex multimodal downstream tasks like modality generation.

\subsection{Experiments and Evaluations}
To comprehensively benchmark MM-OpenFGL, we conduct evaluations across four dimensions: data analysis, effectiveness, Robustness, and efficiency.

\textbf{Data Analysis.}
\ding{182} \underline{\textit{Feature KL Divergence}}: Quantifies feature distribution shifts across clients to capture variations arising from environmental or geographical factors. \ding{183} \underline{\textit{Label-Topology Correlation}}: Utilizes multi-level homophily metrics to analyze the interplay between label distributions and graph structure. \ding{184} \underline{\textit{Topology Statistics}}: Assesses cross-client disparities in local structure (e.g., node degree, centrality), which critically influence GNN performance and collaborative dynamics in distributed settings.

\textbf{Effectiveness.}
We employ standard metrics across tasks: Accuracy, F1, Recall, and Precision for node classification; For modaility-level tasks, AP and AUC-ROC for matching; Recall@K and MRR for retrieval; and BLEU, ROUGE-L, and CIDEr for generation. See Appendix~\ref{appendix: metric} for details.

\textbf{Robustness.}
To assess stability in practical deployment, we evaluate five critical dimensions: noise (simulating data quality issues), sparsity (handling incomplete data due to scarcity or cost), homophily (assessing topological impact), generalization (testing adaptability), and privacy preservation (specifically utilizing differential privacy).

\textbf{Efficiency.}
To facilitate the practical deployment of algorithms, we evaluate the efficiency of current baseline methods from both theoretical and experimental perspectives, providing a comprehensive view of their scalability and deployment feasibility.

\begin{table}[t]
    \caption{Comparison across three categories of baselines on node classification tasks, measured by accuracy (\%).}
    \resizebox{\linewidth}{!}{
    \begin{tabular}{c|ccc|ccc}
        \toprule
         \textbf{Algorithms} & \textbf{Movies} &\textbf{RedditS} & \textbf{Cloth} & \textbf{Bili Music} & \textbf{DY} & \textbf{Ele-fashion} \\
         \midrule
         MMGCN-Local& \makecell{47.88\\{\scriptsize \(\pm\)1.07}}&\makecell{83.67\\{\scriptsize \(\pm\)1.11}} &\makecell{88.20\\{\scriptsize \(\pm\)0.57}} &\makecell{75.82\\{\scriptsize \(\pm\)0.36}} &\makecell{75.17\\{\scriptsize \(\pm\)0.90}} & \makecell{84.52\\{\scriptsize \(\pm\)1.91}}\\
         MGAT-Local&\makecell{53.86\\{\scriptsize \(\pm\)0.17}} &\makecell{90.25\\{\scriptsize \(\pm\)0.49}} &\makecell{92.78\\{\scriptsize \(\pm\)0.13}} &\makecell{\textbf{84.96}\\{\scriptsize \(\pm\)\textbf{0.26}}} &\makecell{\textbf{84.19}\\{\scriptsize \(\pm\)\textbf{0.89}}} &\makecell{\textbf{94.89}\\{\scriptsize \(\pm\)\textbf{0.17}}} \\
         GSMN-Local&\makecell{51.67\\{\scriptsize \(\pm\)0.96}} &\makecell{91.73\\{\scriptsize \(\pm\)0.21}} &\makecell{\textbf{94.28}\\{\scriptsize \(\pm\)\textbf{0.11}}} &\makecell{84.80\\{\scriptsize \(\pm\)0.45}} &\makecell{83.04\\{\scriptsize \(\pm\)0.03}} &\makecell{88.06\\{\scriptsize \(\pm\)1.89}} \\
         \midrule
         PEPSY &\makecell{47.41\\{\scriptsize \(\pm\)0.26}} &\makecell{93.46\\{\scriptsize \(\pm\)0.13}} &\makecell{88.66\\{\scriptsize \(\pm\)0.12}} & \makecell{69.85\\{\scriptsize \(\pm\)0.31}}& \makecell{67.39\\{\scriptsize \(\pm\)1.39}}&\makecell{81.66\\{\scriptsize \(\pm\)1.18}} \\
         FedTGP &\makecell{50.59\\{\scriptsize \(\pm\)0.26}} &\makecell{88.77\\{\scriptsize \(\pm\)0.14}} &\makecell{88.71\\{\scriptsize \(\pm\)0.11}}&\makecell{70.45\\{\scriptsize \(\pm\)1.16}} &\makecell{70.96\\{\scriptsize \(\pm\)1.54}} &\makecell{84.04\\{\scriptsize \(\pm\)0.78}} \\
         FedMVP &\makecell{53.31\\{\scriptsize \(\pm\)0.11}} &\makecell{94.05\\{\scriptsize \(\pm\)0.11}} & \makecell{88.98\\{\scriptsize \(\pm\)0.09}}& \makecell{69.43\\{\scriptsize \(\pm\)2.29}}& \makecell{70.06\\{\scriptsize \(\pm\)4.41}}&\makecell{87.17\\{\scriptsize \(\pm\)0.56}} \\
         FedMAC &\makecell{52.55\\{\scriptsize \(\pm\)0.16}} &\makecell{\textbf{94.17}\\{\scriptsize \(\pm\)\textbf{0.12}}} &\makecell{88.03\\{\scriptsize \(\pm\)0.05}} &\makecell{71.60\\{\scriptsize \(\pm\)0.30}} &\makecell{71.19\\{\scriptsize \(\pm\)3.82}} &\makecell{86.59\\{\scriptsize \(\pm\)1.43}} \\
         \midrule
         FedLap &\makecell{\textbf{55.43}\\{\scriptsize \(\pm\)\textbf{0.27}}} &\makecell{\textbf{94.17}\\{\scriptsize \(\pm\)\textbf{0.01}}} &\makecell{89.02\\{\scriptsize \(\pm\)0.04}} &\makecell{72.75\\{\scriptsize \(\pm\)7.20}} &\makecell{77.45\\{\scriptsize \(\pm\)4.55}} &\makecell{86.76\\{\scriptsize \(\pm\)1.45}} \\
         FedSPA & \makecell{43.86\\{\scriptsize \(\pm\)0.45}}&\makecell{93.02\\{\scriptsize \(\pm\)0.94}} &\makecell{91.24\\{\scriptsize \(\pm\)0.55}} &\makecell{67.63\\{\scriptsize \(\pm\)0.16}} &\makecell{64.00\\{\scriptsize \(\pm\)0.03}} &\makecell{76.95\\{\scriptsize \(\pm\)0.02}} \\
         S2FGL & \makecell{41.58\\{\scriptsize \(\pm\)1.47}}&\makecell{93.28\\{\scriptsize \(\pm\)0.69}} &\makecell{90.82\\{\scriptsize \(\pm\)0.68}} &\makecell{67.80\\{\scriptsize \(\pm\)0.25}} &\makecell{65.48\\{\scriptsize \(\pm\)0.61}} &\makecell{78.48\\{\scriptsize \(\pm\)1.09}} \\
        \bottomrule
    \end{tabular}
    }
    \label{Q2}%
    \vspace{-2mm}
\end{table}

\begin{figure}[t]
  \includegraphics[width=\linewidth]{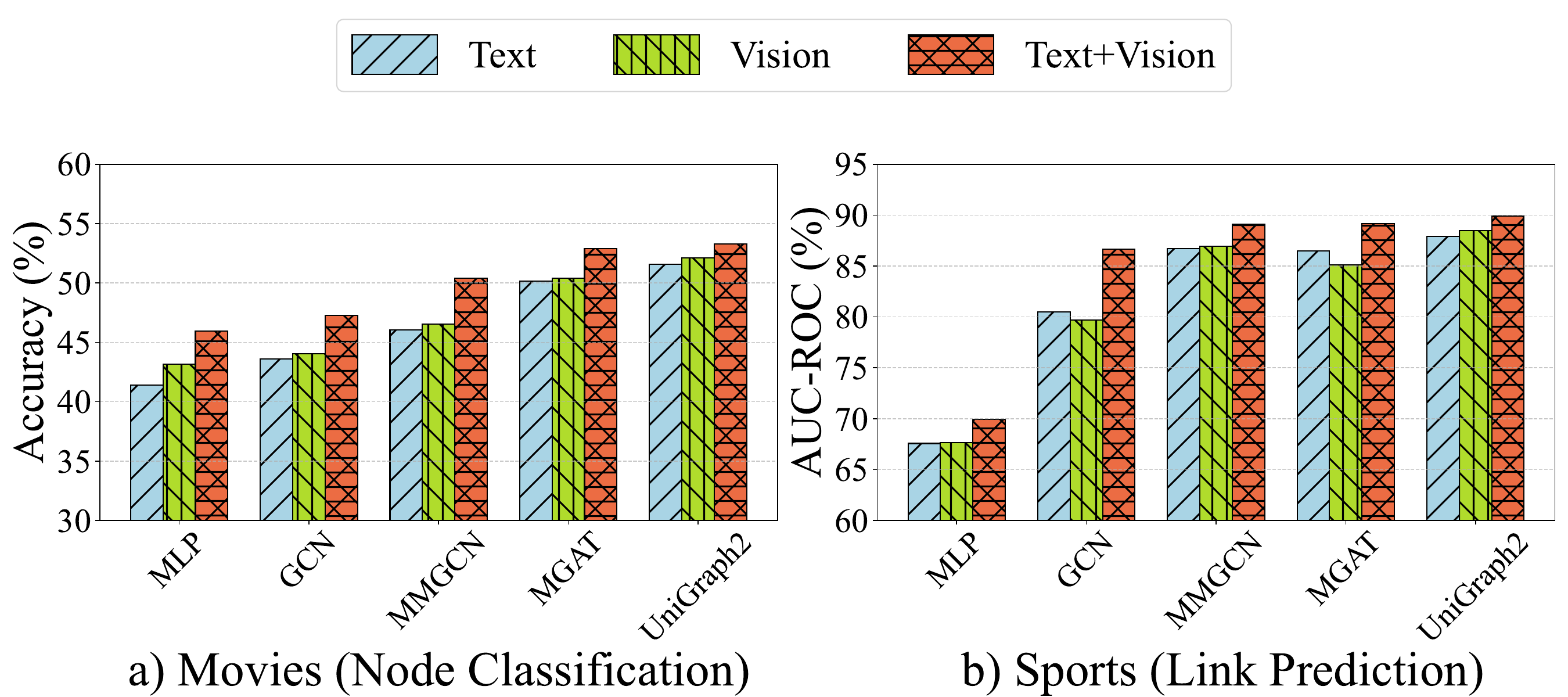}
  \vspace{-5mm}
  \caption{Performance between GNNs using different modalities.}
  \label{fig: Q1}
  \vspace{-2mm}
\end{figure}

\vspace{-1mm}

\section{Experiments and Analysis}
\label{sec: experiments and analysis}
In this section, we systematically investigate the MM-FGL by answering the following 10 research questions:
For \underline{\textit{Necessity}}, \textbf{Q1}: Is it necessary to integrate multimodal semantics and graph topology under federated scenarios? \textbf{Q2}: Can existing federated adaptations effectively handle the heterogeneity of multimodal-attributed graphs? For \underline{\textit{Effectiveness}}, \textbf{Q3}: How do algorithms perform under the proposed simulation scenarios? \textbf{Q4}: How do algorithms generalize across multimodal datasets and multi-level downstream tasks? For \underline{\textit{Robustness}},
\textbf{Q5}: How do heterogeneous FL methods perform under extreme modality-missing scenarios?
\textbf{Q6}: How does the choice of upstream feature encoders influence the robustness of FL algorithms?
\textbf{Q7}: How do existing algorithms perform under local noise and sparse data conditions? For \underline{\textit{Efficiency}},
\textbf{Q8}: What are the theoretical and practical time and memory complexities of the different baselines?
\textbf{Q9}: How do the methods compare in terms of communication overhead?
\textbf{Q10}: How efficient are the algorithms with respect to convergence speed?

\subsection{Necessity Analysis (Answer for \textbf{Q1} and \textbf{Q2})}
\label{sec: exp_necessity}

To investigate \textbf{Q1}, we compare single-modality baselines with multimodal architectures across datasets of varying scales and tasks. Specifically, we adopt MLP and GCN as representative single-modality models, and MM-GCN, MGAT, and UniGraph as representative multimodal GNNs (MM-GNNs). As shown in Fig.~\ref{fig: Q1}, our observations can be summarized as follows: \ding{182} \underline{\textit{Multimodal Advantage:}} MM-GNNs consistently and substantially outperform single-modality GNNs across all settings, indicating that multimodal semantics provide complementary signals that alleviate modality-specific ambiguities and lead to more informative node representations. \ding{183} \underline{\textit{Structural Necessity:}} Among single-modality baselines, GCN significantly outperforms MLP. Since MLP operates solely on node attributes whereas GCN leverages neighborhood aggregation, this gap underscores the essential role of graph topology in propagating label information and smoothing feature representations.

\textbf{Our Conclusion (C1).} Neither graph structure nor multimodal semantics alone suffices; their integration is essential to achieve optimal performance. 

To address \textbf{Q2}, we benchmark three categories of methods: isolated baselines, where MM-GNNs such as MM-GCN and MGAT are trained locally on each client without communication; multimodal FL adapted to graphs, including FedMVP and FedMAC; and FGL adapted to multimodal data, including FedLap and S2FGL. We evaluate the above methods on node classification (Movies, RedditS and Cloth) and link prediction (Bili Music, DY and Ele-fashion). As reported in Table \ref{Q2}, our observations can be summarized as follows: \ding{182} \underline{\textit{Limited Gains.}} Naively adapted federated methods outperform isolated training on only two datasets, Movies and RedditS, but perform worse on the remaining four datasets: Cloth, Bili, Music, and DY. \ding{183} \underline{\textit{No Clear Winner.}} There is no statistically significant difference or consistent advantage between adapting vision-oriented FL methods to graphs and adapting FGL methods to multimodal data.

\textbf{Our Conclusion (C2).} Naive adaptations of existing strategies prove ineffective in MM-FGL settings, as they neither align conflicting multimodal signals nor resolve structural mismatches across clients. 
Consequently, federated collaboration can underperform isolated training, underscoring the need for methods specifically designed for MM-FGL.

\vspace{0.1cm}
\subsection{Performance Comparison (Answer for \textbf{Q3} and \textbf{Q4})}

To better simulate real-world conditions, we design eight simulation settings by jointly varying three orthogonal factors: 
\ding{182} modality distribution (IID vs. NonIID), 
\ding{183} topology availability (available vs. unavailable), and 
\ding{184} label distribution (IID vs. NonIID). 
These settings span from an \underline{\textit{ideal case}} (i.e., all modalities and topology are available with IID labels) to the \underline{\textit{most challenging case}} (i.e., clients possess disjoint modalities, lack topology information, and exhibit label-NonIID distributions).

\begin{figure}[t]
  \includegraphics[width=\linewidth]{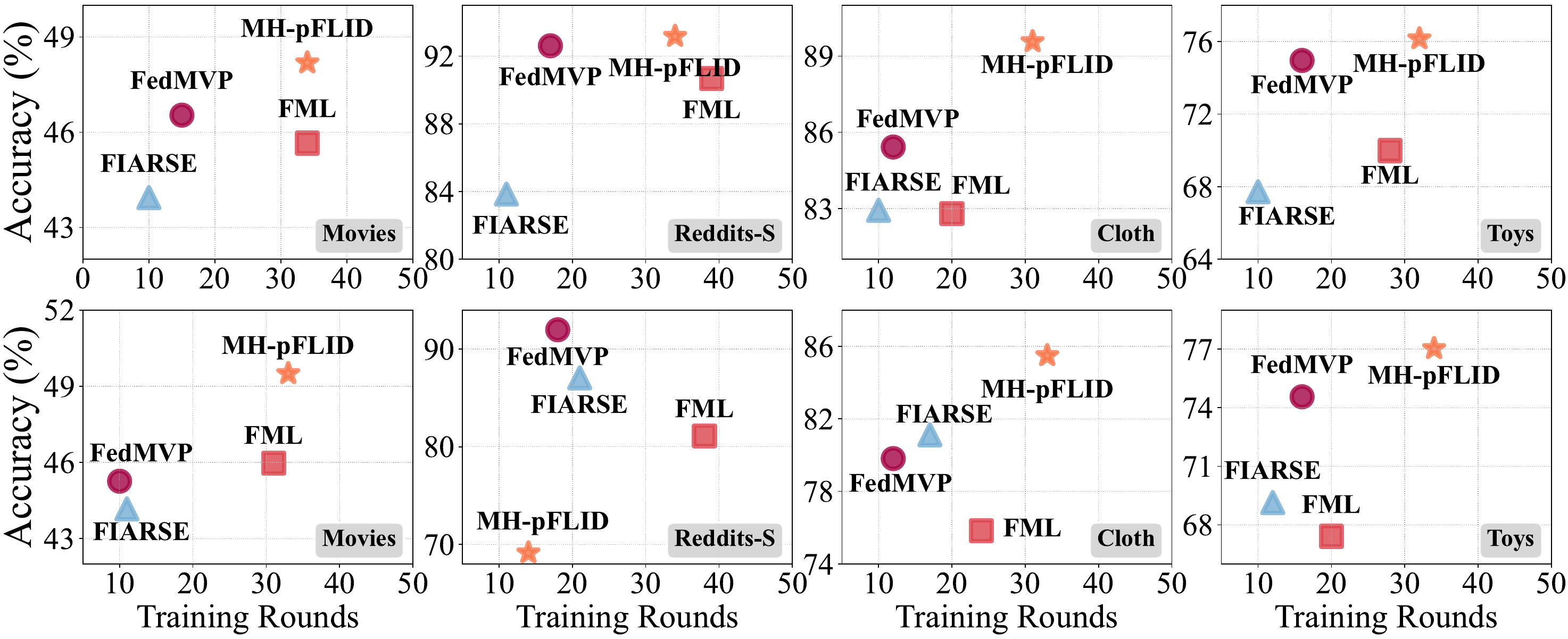}
  \vspace{-0.5cm}
  \caption{Best-round node classification performance for different simulations. \textbf{Top}: modality-NonIID, topology unavailable, label-IID; \textbf{Bottom}: modality-NonIID, topology available, label-NonIID.}
  \vspace{-2mm}
  \label{fig: Q3}
  
\end{figure}

To answer \textbf{Q3}, we evaluate existing methods across all eight settings, with particular emphasis on modality-NonIID scenarios. 
Due to the inherent difficulty of handling missing or disjoint modalities, we focus on heterogeneous FL methods (FedMVP, FIARSE, FML, and MH-pFLID), which constitute the core and most realistic challenge captured by our benchmark. As shown in Fig.~\ref{fig: Q3}, we observe consistent patterns across seven datasets:
\ding{182} \underline{\textit{MH-pFLID Dominance:}} MH-pFLID consistently achieves the largest performance region across nearly all datasets, demonstrating strong robustness even in high-variance domains such as Grocery and Bili Dance;
\ding{183} \underline{\textit{Dataset Sensitivity:}} Competing methods exhibit substantial performance fluctuations across datasets (e.g.,  FML performs competitively on Cloth but degrades markedly on Grocery), indicating limited stability under heterogeneous data distributions;
\ding{184} \underline{\textit{Impact of NonIID:}} While the relative ranking of methods remains largely unchanged when moving from label-IID to label-NonIID settings, the overall performance consistently deteriorates, confirming that modality and label-NonIID pose significantly greater challenges to model representation.

\textbf{Our Conclusion (C3).} Modality-NonIID coupled with label-NonIID significantly degrades the performance of general baselines, whereas specialized heterogeneous FGL methods like MH-pFLID demonstrate consistent robustness.

\begin{table*}[t]
    \centering
    \caption{Performance comparison. The highest results are highlighted in \textbf{bold}, while the second-highest results are marked with \underline{underline}. }
    \label{q4:performance Comparison}
    \resizebox{\textwidth}{!}{
    \begin{tabular}{c|cc|cc|cc|cc}
    \toprule
 & \multicolumn{2}{c|}{\textbf{Node Classification (ACC)}} & \multicolumn{2}{c|}{\textbf{Modality Match. (AUC-ROC)}} & \multicolumn{2}{c|}{\textbf{Modality Retrieval (Recall@5)}} & \multicolumn{2}{c}{\textbf{G2Text (ROUGE-L)}} \\
 
 \cline{2-9}
 \noalign{\smallskip}
\multirow{-2}{*}{\textbf{Methods}} & \textbf{Toys} & \textbf{Grocery} & \textbf{KU} & \textbf{Bili Food} & \textbf{QB} & \textbf{Bili Cartoon} & \textbf{Flickr30k} & \textbf{MultiMET Ads} \\
    \midrule
         Fed-MGNet&79.20 ± 0.32 & 69.61 ± 0.81& \underline{57.61} ± 0.49 & 57.43 ± 1.00& 84.46 ± 4.26 & 67.08 ± 3.90 & 16.24 ± 0.22 & 16.15 ± 0.35 \\
         Fed-MHGAT&79.03 ± 0.13 &71.54 ± 0.89 &55.90 ± 1.39 &57.90 ± 1.78 &87.98 ± 5.76 &71.52 ± 4.35 & 17.02 ± 0.16& 15.44 ± 0.18 \\
         Fed-RevGAT&79.19 ± 0.06 &69.25 ± 3.35 &56.90 ± 1.40 & \textbf{59.08} ± 0.44 &82.94 ± 2.14 & 70.24 ± 2.24& 18.22 ± 0.27 & 17.45 ± 0.30 \\
         \midrule
         FML&79.08 ± 0.29 &70.48 ± 1.46 &56.59 ± 0.43 &57.31 ± 0.74 &89.66 ± 0.18 &72.31 ± 3.49& 19.05 ± 0.11 & 18.42  ± 0.28 \\
         FedMVP&78.98 ± 0.07 &70.59 ± 1.80 & \textbf{58.02} ± 0.47 & \underline{58.97} ± 0.99 & 90.32 ± 1.67 &70.28 ± 1.07& 18.53 ± 0.30 & 16.71 ± 0.09 \\
         S2FGL&\textbf{79.91} ± 0.45 &69.41 ± 2.54 &55.89 ± 1.40 &57.90 ± 1.59 &91.23 ± 2.23 & \underline{78.30} ± 5.27& 17.30 ± 0.21 & 17.88 ± 0.15 \\
         FedLap&79.07 ± 0.20 &69.87 ± 0.19 &56.42 ± 0.85 &56.36 ± 1.40 &\underline{92.49} ± 3.43 &74.47 ± 5.54& 16.95 ± 0.20 & 17.22 ± 0.22 \\
         FedSPA&79.11 ± 0.07 &69.10 ± 3.46 &56.47 ± 2.19 &57.67 ± 0.89 &83.61 ± 4.82 &75.93 ± 4.76& 18.40 ± 0.25 & 18.32 ± 0.33 \\
         FedIIH&79.08 ± 0.12 &69.83 ± 2.26 &57.36 ± 0.56 &57.69 ± 2.81 &90.07 ± 4.20 &76.78 ± 3.73& 17.47 ± 0.14 & 16.48 ± 0.26 \\
         FedSSP&79.14 ± 0.15 & \underline{71.65} ± 0.98 &56.72 ± 0.54 &58.96 ± 0.98 & 90.96  ± 3.55 & 78.09 ± 4.46& 19.22 ± 0.22 & 18.37 ± 0.30 \\
         \midrule
         GFT&79.27 ± 0.07 &69.41 ± 0.65 &55.80 ± 0.80 &57.68 ± 1.01 &85.87 ± 4.34 &77.57 ± 2.77 & 18.36 ± 0.31 & 17.94 ± 0.28 \\
         OFA&79.13 ± 0.25 &70.20 ± 1.45 &55.93 ± 0.68 &56.91 ± 0.73 & 77.31 ± 3.29 &66.87 ± 3.35  &  16.74 ± 0.12&15.85 ± 0.06 \\
         GraphCLIP& \underline{79.35} ± 0.09 & \textbf{72.07} ± 0.83 &55.63 ± 0.32 &58.04 ± 1.27 &\textbf{94.89} ± 3.32 &\textbf{83.12} ± 2.07  & \textbf{20.65 ± 0.16} & \textbf{19.23 ± 0.07} \\
         UniGraph2&79.11 ± 0.12 & 69.17 ± 2.27 &56.87 ± 0.75 &57.76 ± 1.43 &87.18 ± 3.96 &68.35 ± 4.06   & \underline{20.14 ± 0.20} & \underline{19.06 ± 0.13} \\
         \bottomrule
    \end{tabular}
     }
     \vspace{-0.1cm}
\end{table*}

\begin{figure}[t]
  \includegraphics[width=\linewidth]{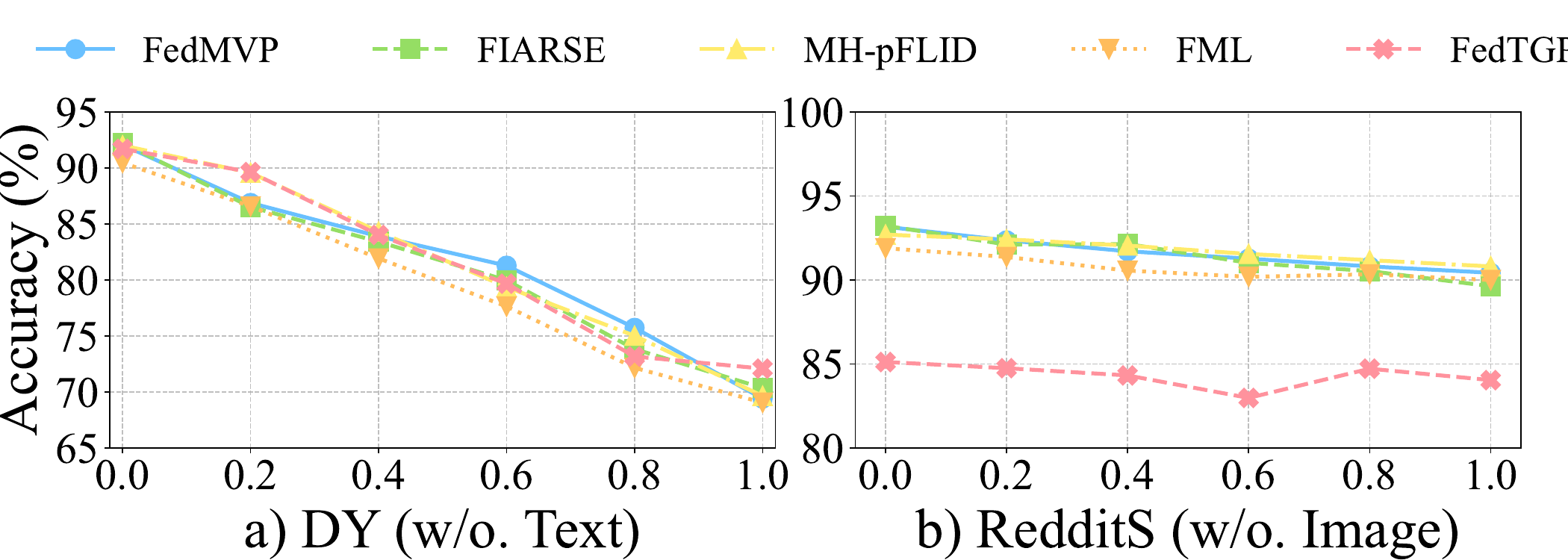}
  \vspace{-6mm}
  \caption{Algorithm robustness under modality missing on DY and RedditS datasets (node classification task).}
  \vspace{-4mm}
  \label{fig: Q5}
\end{figure}

\begin{figure}[t]
  \includegraphics[width=\linewidth]{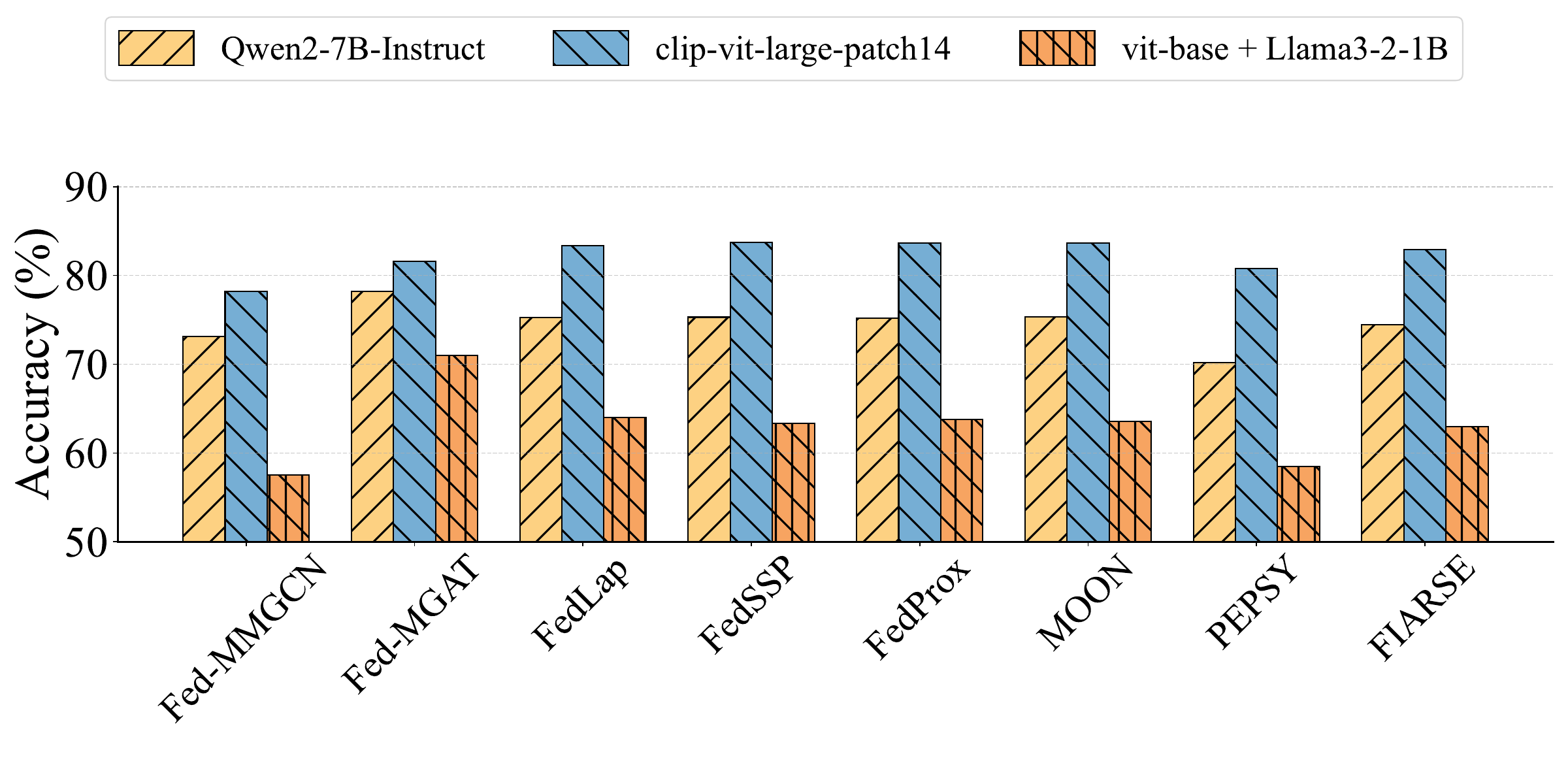}
  \vspace{-1cm}
  \caption{Algorithm robustness under different feature encoders for the Toys dataset (node classification task).}
  \label{fig: Q6}
  \vspace{-0.3cm}
\end{figure}

\begin{figure*}[t]
  \includegraphics[width=\textwidth]{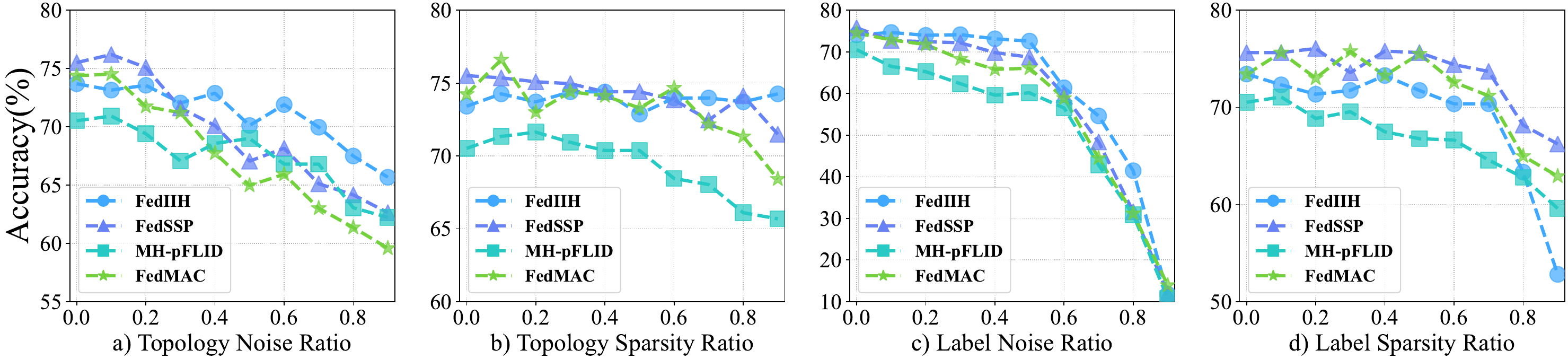}
  \vspace{-6mm}
  \caption{Robustness performance comparisons under topology/ label noise and sparsity on the Bili Movie dataset (node classification).}
  \label{fig: Q7}
  \vspace{-4mm}
\end{figure*}

To address \textbf{Q4}, we conducted a systematic evaluation of the four categories of included methods across different downstream tasks. Specifically, we selected the most representative works from each category. For multimodal GNNs, we chose MGNet, MHGAT, and RevGAT, and implemented their federated versions. For federated learning algorithms, we selected methods such as FML, FedMVP, and S2FGL, covering both standard FL and heterogeneous FL approaches. For graph foundation models, we included GFT, OFA, GraphCLIP, and UniGraph2. These methods were evaluated on node classification (Toys and Grocery), modality matching (KU and Bili‑Food), modal retrieval (QB and Bili‑Cartoon) and modality generation (Flickr30k and MultiMET Ads). The results are summarized in the Table~\ref{q4:performance Comparison}. From the Table~\ref{q4:performance Comparison}, we can observe that:
\ding{182} \underline{\textit{GFMs Perform Better}}: Compared to multimodal graph learning models and multimodal federated learning algorithms, graph foundation models demonstrate superior performance on a wider range of downstream tasks.
\ding{183} \underline{\textit{GFMs are More Scalable}}: GFMs can be adapted to a wider range of downstream tasks, while other methods cannot, such as modality generation tasks.

\textbf{Our Conclusion (C4)}: 
GFMs consistently outperform traditional MM-GNNs and FL baselines across diverse tasks, demonstrating superior generalization and scalability. Uniquely, GFMs possess the versatility to handle complex modality-level tasks, such as modality generation, which remain beyond the capabilities of task-specific architectures.

\subsection{Robustness Analysis (Answer for \textbf{Q5}, \textbf{Q6} and \textbf{Q7})}
To answer \textbf{Q5}, we assessed the robustness of heterogeneous FL methods by subjecting them to a controlled gradient of modality missing. Specifically, in the label-IID scenario, we varied the modality missing rate from 0.0 (complete modalities) to 1.0 (complete miss of a specific modality) on the DY and RedditS datasets. We selected five representative heterogeneous FL methods, including FedMVP, FIARSE, MH-pFLID, FML, and FedTGP.

As shown in Fig.~\ref{fig: Q5}, we observe the following patterns:
\ding{182} \underline{\textit{Degradation under Modality Missing:}} 
All methods exhibit a consistent decline in accuracy as the modality missing rate increases, confirming that the loss of modality information fundamentally weakens the discriminative capability of client models;
\ding{183} \underline{\textit{Dataset-dependent Sensitivity:}} 
The severity of performance degradation varies substantially across datasets. For example, DY experiences a sharp accuracy drop of over 20\% as the missing rate approaches 1.0, indicating a strong reliance on cross-modal complementarity. In contrast, RedditS remains highly stable, with leading methods (e.g., FedMVP and MH-pFLID) maintaining over 90\% accuracy even under extreme missing rates, suggesting that the remaining modality provides sufficient semantic redundancy;
\ding{184} \underline{\textit{Method Robustness and Inconsistency:}} 
FedMVP and MH-pFLID demonstrate the most consistent robustness across datasets and effectively mitigate severe performance degradation. 
In contrast, FedTGP exhibits unstable behavior. It performs competitively on DY but underperforms on RedditS across all modality missing rates, reflecting limited adaptability to dominant-modality settings.

\textbf{Our Conclusion (C5).} The robustness of heterogeneous FL methods to modality missing varies substantially across datasets. 
Methods remain stable on datasets with sufficient semantic redundancy, but exhibit pronounced degradation on tasks that rely on strong cross-modal coupling.

To solve \textbf{Q6}, we evaluate the performance of different methods under data feature extracted by different encoders on the same dataset. As illustrated in Fig~\ref{fig: Q6}, the following observations were obtained:
\ding{182} \underline{\textit{Feature Encoder Sensitivity}}: All methods exhibit substantial variations in performance across different feature encoders, indicating a strong dependence of existing approaches on the initial data representations.
\ding{183} \underline{\textit{Direction for Improvement}}: Among the compared methods, Fed-MGAT demonstrates the strongest robustness to initial features (showing the smallest performance variance across different feature encoders), suggesting that the robustness of current multimodal federated graph learning methods to initial feature quality still leaves room for improvement.

\textbf{Our Conclusion (C6)}. Given the proliferation of multimodal feature encoders, enhancing robustness to data feature represents a promising research direction.

\begin{table*}[t]
    \caption{Comprehensive efficiency analysis on the Bili Movie dataset for node classification. We evaluate performance across theoretical complexity, empirical time/memory costs, communication overhead, and convergence speed.}
    \vspace{0.05cm}
    \centering
    \label{tab: complexity}
    \resizebox{\textwidth}{!}{
    \begin{tabular}{c|c|cc|cc|cc}
    \toprule
         \textbf{Method} & \textbf{Accuracy (\%)} & \textbf{Theo. Time} & \textbf{Prac. Time} & \textbf{Theo. Mem} & \textbf{Prac. Mem} & \textbf{Com.} & \textbf{Conver. ($\downarrow$)}  \\
    \midrule
         FML&92.82 ± 0.10 & $O(Kmf+nf^2)$ &35.13s & $O(N(f^2+F^2))$&1637K &24.28K & 10\\
         FIARSE& 94.01 ± 0.13&$O(Kmf+nf^2)$ &24.68s &$O(Nf^2)$ &579K &102K & 5\\
         MH-pFLID&93.96 ± 0.01 & $O(Kmf+nf^2)$&39.28s & $O(N(f^2+F^2))$&2161K &745K & 7\\
         PEPSY&93.37 ± 0.18 &$O(Kmf+nf^2)$ &52.13s & $O(N(f^2+F^2))$&1069K &39.5K & 25\\
         FedMVP&94.03 ± 0.23 &$O(Kmf+nf^2)$ &25.22s &$O(Nf^2)$ &54.6K &18.2K & 4\\
         FedMAC&93.89 ± 0.25 & $O(Kmf+nf^2)$&30.32s &$O(Nf^2+hc)$ &292K &42.1K & 5\\
         FedSPA&93.87 ± 0.42 & $O(Kmf+nf^2+En^2)$&1145.1s &$O(N(f^2+n^2))$ &86.4M &120K & 30\\
         FedIIH&93.45 ± 0.29 & $O(Kmf+nf^2+n^2)$&273.21s &$O(Nf^2+n^2)$ &1012K &24.5K & 34\\
         FedSSP&94.17 ± 0.06 & $O(Kmf+nf^2)$&28.67s & $O(Nf^2)$&64.8K &26.7K & 4\\
         S2FGL&93.41 ± 0.12 &$O(Kmf+nf^2+qn^2)$     &399.61s &$O(Nf^2+n^2+qn)$ &935K &112K & 45\\
         FedLap&94.16 ± 0.11 & $O(Kmf+nf^2)$&26.74s & $O(Nf^2+n^2)$&411K &31.4K & 4\\
    \bottomrule
    \end{tabular}
    }
    \vspace{+0.05cm}
\end{table*}

To answer \textbf{Q7}, we evaluated the stability of algorithms under four distinct perturbation settings: topology noise, topology sparsity, label noise, and label sparsity. We varied the perturbation ratio from 0.0 to 0.9 to stress-test the methods.
As illustrated in the Fig.~\ref{fig: Q7}, we observe that robustness is highly asymmetric across perturbation types: \ding{182} \underline{\textit{Robustness to Label Perturbations:}}
Robustness is highly asymmetric across perturbation types. Once label noise exceeds 0.5, all methods experience a sharp performance collapse, and extreme label sparsity leads to similarly severe degradation, with FedIIH suffering the most; \ding{183} \underline{\textit{Robustness to Topological Perturbations:}}
Most methods remain stable under topology noise and topology sparsity, with substantially milder performance degradation than under label perturbations. However, MH-pFLID shows pronounced sensitivity to structural loss.

\textbf{Our Conclusion (C7).} In MMFGL settings, rich node semantics can effectively compensate for severe topological loss, rendering models relatively robust to structural sparsity. However, label integrity remains the single most critical factor, as current algorithms lack mechanisms to withstand high-ratio label noise.
\subsection{Efficiency Analysis}

To answer \textbf{Q8}, we provide the theoretical complexity of representative methods, and the results are shown in Table~\ref{tab: complexity}. The detailed complexity analysis is provided in Appendix \ref{appendix: complexity}. In addition, we evaluated the practical complexity of the methods through experiments, and our observations are as follows:
\ding{182} \underline{\textit{Time/ Memory Bottleneck}}: In multimodal federated graph learning settings, the true computational bottleneck lies not in the number of training rounds, but rather in the introduction of a dense adjacency matrix. 
\ding{183} \underline{\textit{Server-side Computational Cost}}: The computational cost of the similarity matrix or personalized aggregation on the server side is dominated by the number of clients. Even though clients can process in parallel in methods such as MH-pFLID, PEPSY, and FedSPA, the server becomes the bottleneck, which explains why some approaches exhibit a prolonged overall training time despite moderate client-side computational load in empirical evaluations.

\textbf{Our Conclusion (C8).} Optimization should prioritize sparse operations and decentralized aggregation over reducing communication rounds, lowering time and memory overhead while mitigating server-side bottlenecks.

To answer \textbf{Q9}, we calculated the communication overhead for each round of existing federated algorithms. As shown in the Table~\ref{tab: complexity}, the communication overhead of MH-pFLID is significantly higher than other methods, indicating a major weakness. In contrast, FedMVP, FML, FedIIH and FedSSP have advantages in terms of communication overhead.

\textbf{Our Conclusion} (\textbf{C9}).  Methods that transmit compact semantic summaries (e.g., prototypes) rather than full model parameters significantly reduce communication overhead while maintaining high accuracy through lightweight alignment mechanisms.

To address \textbf{Q10}, we evaluate convergence rates over 50 federated rounds. The convergence round, reported in Table~\ref{tab: complexity}, is defined as the minimum number of rounds required to reach 99.5\% of the best achieved performance, where smaller values indicate faster convergence. Our observations are as follows:
\ding{182} \underline{\textit{Significant Inter-Method Disparities:}} Convergence results reveal a sharp dichotomy. Lightweight semantic alignment methods, such as FedMVP, FedSSP, and FedLap, converge in 4 rounds with top accuracy, while methods based on complex structural reconstruction or dense aggregation, like S2FGL (45 rounds) and FedIIH (34 rounds), require nearly ten times more rounds.
\ding{183} \underline{\textit{Impact of Design Choices:}} Convergence speed is highly sensitive to aggregation strategies and prompt/token utilization. Efficient designs accelerate training and reduce communication overhead, offering practical guidance for federated graph learning.

\textbf{Our Conclusion} (\textbf{C10}). Convergence speed is highly sensitive to aggregation strategies and prompt/token utilization. Efficient designs accelerate training and reduce communication overhead, offering practical guidance for FGL.

\vspace{0.3cm}
\section{Conclusion and Future Directions}

In this work, we introduce MM-OpenFGL, a comprehensive benchmark framework that, for the first time, systematically formalizes and advances the emerging paradigm of multimodal federated graph learning (MMFGL).
By integrating 19 heterogeneous datasets, 57 state-of-the-art algorithms, and a tri-dimensional simulation protocol that instantiates eight representative federated scenarios, MM-OpenFGL establishes a rigorous and unified evaluation standard for assessing necessity, effectiveness, robustness, and efficiency across diverse settings.
Extensive experiments distill ten key insights, revealing the limitations of existing forced adaptation strategies while underscoring the critical roles of cross-modal alignment and structural resilience.
Together with a modular open-source library, MM-OpenFGL lowers entry barriers, enables fair comparison, and accelerates research in privacy-preserving multimodal graph intelligence.

\section*{Impact Statement}

This paper presents work whose goal is to advance the field of 
Machine Learning. There are many potential societal consequences 
of our work, none which we feel must be specifically highlighted here.

\bibliography{example_paper}
\bibliographystyle{icml2025}

\newpage
\appendix
\onecolumn
\section{Appendix}

\subsection{Datasets}

\begin{table*}[t]
\centering
\caption{Statistical information of the NineRec downstream datasets.}
\label{tab:dataset_stats1} 

\setlength{\tabcolsep}{0pt}          
\renewcommand{\arraystretch}{1.2}    
\setlength{\extrarowheight}{2pt}     
\footnotesize

\begin{tabular*}{\textwidth}{
  @{\extracolsep{\fill}}            
  >{\centering\arraybackslash}m{1.6cm}  
  >{\centering\arraybackslash}m{1.2cm}  
  >{\centering\arraybackslash}m{1.4cm}  
  >{\centering\arraybackslash}m{1.8cm}  
  >{\centering\arraybackslash}m{1.6cm}  
  >{\centering\arraybackslash}m{1.2cm}  
  >{\centering\arraybackslash}m{1.5cm}  
  >{\centering\arraybackslash}m{1.8cm}  
  @{}                               
}
\toprule
Dataset & Users & Items & Interactions & Density($10^{-3}$) & TextRatio & Split & Description \\
\midrule
Bili Cartoon & 4{,}724  & 30{,}300 & 215{,}443 & 1.505 & 0.1559 & 60\%/20\%/20\% & Cart Videos \\
Bili Dance   & 2{,}307  & 10{,}715 & 83{,}392  & 3.374 & 0.2153 & 60\%/20\%/20\% & Dance Videos \\
Bili Food    & 1{,}579  & 6{,}549  & 39{,}740  & 3.843 & 0.2411 & 60\%/20\%/20\% & Food Videos \\
Bili Movie   & 3{,}509  & 16{,}525 & 115{,}576 & 1.993 & 0.2123 & 60\%/20\%/20\% & Movie Videos \\
Bili Music   & 6{,}038  & 50{,}664 & 360{,}177 & 1.177 & 0.1192 & 60\%/20\%/20\% & Music Videos \\
DY            & 8{,}299  & 20{,}398 & 139{,}834 & 0.826 & 0.4069 & 60\%/20\%/20\% & Short Videos \\
KU            & 5{,}370  & 2{,}034  & 18{,}519  & 1.695 & 1.0000 & 60\%/20\%/20\% & Knowl Videos \\
QB            & 17{,}722 & 6{,}121  & 133{,}664 & 1.232 & 1.0000 & 60\%/20\%/20\% & Q\&A Videos \\
TN            & 20{,}211 & 3{,}334  & 122{,}576 & 1.819 & 1.0000 & 60\%/20\%/20\% & News Videos \\
PixelRec50K   & 50{,}000 & 82{,}865 & 989{,}494 & 0.239 & 0.7615 & 60\%/20\%/20\% & Pixel Vid Rec. \\
\bottomrule
\end{tabular*}
\end{table*}

\begin{table*}[t]
\centering
\caption{Statistical information of the GLAMI-1M and MultiMET datasets.}
\label{tab:glami_multimet_stats}

\setlength{\tabcolsep}{0pt}       
\renewcommand{\arraystretch}{1.2} 
\setlength{\extrarowheight}{2pt}  
\footnotesize

\begin{tabular*}{\textwidth}{
  @{\extracolsep{\fill}}  
  >{\centering\arraybackslash}m{1.8cm}  
  >{\centering\arraybackslash}m{2.0cm}  
  >{\centering\arraybackslash}m{1.8cm}  
  >{\centering\arraybackslash}m{1.8cm}  
  >{\centering\arraybackslash}m{1.2cm}  
  >{\centering\arraybackslash}m{1.2cm}  
  >{\centering\arraybackslash}m{1.8cm}  
  @{}  
}
\toprule
Dataset & Items & Images & Texts & Languages & TextRatio & Description \\
\midrule
GLAMI-1M         & 1{,}108{,}532 & 968{,}423 & 734{,}394 & 13 & 1.00 & Fashion \\
MultiM-Ads     & 4{,}064       & 4{,}064   & 3{,}418   & 3  & 0.84 & Ads \\
MultiM-Fb& 1{,}874       & 1{,}874   & 1{,}875   & 4  & 1.00 & Fb posts \\
MultiM-Twit & 4{,}498       & 4{,}498   & 4{,}497   & 5  & 1.00 & Twit posts \\
\bottomrule
\end{tabular*}
\end{table*}

\begin{table*}[t]
\centering
\caption{\textbf{Overview and statistics of MM-Graph and MAGB subsets. 
All datasets provide multimodal node features (image + text).}}

\setlength{\tabcolsep}{0pt}       
\renewcommand{\arraystretch}{1.2} 
\footnotesize

\begin{tabular*}{\textwidth}{
  @{\extracolsep{\fill}}  
  >{\centering\arraybackslash}m{1.6cm}  
  >{\centering\arraybackslash}m{1.2cm}  
  >{\centering\arraybackslash}m{1.0cm}  
  >{\centering\arraybackslash}m{1.4cm}  
  >{\centering\arraybackslash}m{0.9cm}  
  >{\centering\arraybackslash}m{1.2cm}  
  >{\centering\arraybackslash}m{1.1cm}  
  >{\centering\arraybackslash}m{0.9cm}  
  @{}  
}
\toprule
\textbf{Dataset} & \textbf{Task} & \textbf{Modality} & \textbf{Graph} & \textbf{Scale} & \textbf{Domain} & \textbf{Nodes} & \textbf{Classes} \\
\midrule
\multicolumn{8}{c}{\textbf{MM-Graph Subsets}} \\
\midrule

Cloth 
& NodeCls. & Img.+Text & Co-purch. & Small  & Clothing & 232{,}290 & 15 \\

Ele-fashion
& NodeCls. & Img.+Text & Fash prod. & Medium & Fash & 361{,}566 & 12 \\

Sports
& NodeCls. & Img.+Text & Co-purch. & Small & Sports & 173{,}634 & 18 \\

\midrule
\multicolumn{8}{c}{\textbf{MAGB Subsets}} \\
\midrule

Grocery
& NodeCls. & Img.+Text & Co-purch. & Medium & Grocery & 17{,}074 & 20 \\

Movies
& NodeCls. & Img.+Text & Co-purch. & Medium & Movies & 16{,}672 & 20 \\

RedditS
& NodeCls. & Img.+Text & Co-post. & Small & Reddit & 15{,}894 & 20 \\

Toys
& NodeCls. & Img.+Text & Co-purch. & Medium & Toys & 20{,}695 & 18 \\

\bottomrule
\end{tabular*}

\label{tab:mmgraph_magb_overview}
\end{table*}

Our experiments cover three complementary dataset families that reflect the major application settings of our study, including sequence recommendation, multimodal product understanding, and multimodal graph learning. These datasets span video platforms, e-commerce sites, social media, and vision-language corpora, thereby providing a broad evaluation environment for cross-domain and cross-modal generalization.

In graph-based representations used throughout our experiments, a node typically corresponds to a video, product, user-generated post, image region, or medical finding, depending on the dataset. Node features combine textual, visual, or numeric attributes, and task labels when available. For most GNN-based models, these components are concatenated as the input node representation. For vision and video datasets, superpixel or region-based graph construction is used in some cases, where image or frame regions serve as nodes and spatial or semantic relationships define edges. This representation preserves structural cues and supports fine-grained multimodal reasoning.

Below we summarize the datasets used in our experiments.

\textbf{Sequence Recommendation Datasets (NineRec Series)}

\textbf{Bili Cartoon.}~\cite{bili}
Sourced from Bilibili, this subset focuses on animation-related content
and comprises roughly 8k videos. The nodes encode bilingual titles,
descriptions, and tag information, together with visual features
extracted from cover images. User-video interactions form the edge
set. Owing to its focused domain, the dataset is frequently employed
to examine cross-category transfer involving animation content.

\textbf{Bili Dance.}~\cite{bili}
This collection contains around 7.5k dance-performance videos. Alongside
style descriptors and performer metadata, each node includes textual and
visual attributes. Its standardized format within the NineRec suite makes
it suitable for multi-domain recommendation studies.

\textbf{Bili Food.}~\cite{bili}
Approximately 9k videos centered on cooking tutorials and restaurant
explorations form this dataset. Food-category indicators, textual
descriptions, and image-based features constitute the node attributes.
Owing to its distinct thematic structure, the dataset is often used to
evaluate domain adaptation behavior.

\textbf{Bili Movie.}~\cite{bili}
About 8.8k film-related videos, including clips, reviews, and commentary, are included.
Metadata such as film genres and release information, together with cover-image features, are encoded as node attributes.
This dataset enables the investigation of cross-domain recommendation for cinematic content.

\textbf{Bili Music.}~\cite{bili}
This music-focused subset consists of roughly 8.2k items, ranging from music
videos to live performances. Each node records descriptive text, performer
information, and cover-image features, supporting studies on cross-style and
cross-genre recommendation.

\textbf{DY.}~\cite{bili}
Representing short videos from Douyin/TikTok, this 9.5k-item dataset captures
the highly dynamic and lifestyle-oriented nature of short-form media. Hashtags,
captions, and frame-level visual descriptors are included as node features,
while user interactions specify edges. The dataset serves as a natural test of
cross-platform generalizability.

\textbf{KU.}~\cite{bili}
The KU subset, drawn from Kuaishou, contains around 7k knowledge-oriented
videos. Topic labels, explanatory text, and visual content are encoded at the
node level, making the dataset appropriate for studying the interplay between
educational and entertainment content in recommendation tasks.

\textbf{QB.}~\cite{bili}
Originating from Zhihu, this 6.5k-item dataset comprises Q\&A-style explanatory
videos. Category labels, descriptive text, and visual cues form the node
attributes. Its high degree of knowledge specificity provides a distinctive
challenge for cross-domain transfer.

\textbf{TN.}~\cite{bili}
This Tencent News subset consists of approximately 7.8k news and documentary
videos. Attributes include timestamps, category labels, and textual abstracts,
supplemented by cover-image features. It is well suited for domain transfer
involving news-oriented content.

\textbf{Multimodal Product Graph Datasets}

\textbf{Cloth.}~\cite{Zhu2024MosaicOM}
Containing roughly 50k fashion items from Amazon, this dataset represents each product with textual descriptors such as style, material, and brand, together with high-resolution image features.
Graph edges capture complementary relations and co-purchase behavior.

\textbf{Movies.}~\cite{Amazon2018}
This Amazon/MAGB subset, comprising around 45k movie-related products,
encodes film metadata and cover images at the node level. Edges represent
relations such as series membership or co-purchase tendencies.

\textbf{Ele-fashion.}~\cite{Zhu2024MosaicOM}
About 52k wearable-electronics and fashion-related products constitute this
dataset. Nodes blend product specifications, descriptive text, and image
features, while edges reflect complementary or thematic relationships.

\textbf{Grocery.}~\cite{Amazon2018}
This 48k-item Amazon/MAGB subset covers packaged food and household supplies.
Nodes carry packaging descriptions, ingredients, or usage instructions in
textual form, along with product images. Edges indicate co-purchase or
complementary usage patterns.

\textbf{Sports.}~\cite{Zhu2024MosaicOM}
With close to 55k sports-related items, this dataset includes gear, equipment,
and outdoor products. Specification text and image features form the node
attributes; edges reflect typical complementary relationships (e.g., racket and
shuttlecock).

\textbf{Toys.}~\cite{Amazon2018}
As a core component of MAGB, the Toys dataset comprises roughly 42k products.
Age suitability, safety information, materials, and functional descriptions are encoded alongside visual features.
Relations among products, including series membership and accessory compatibility, form the graph edges.

\textbf{PixelRec50K.}~\cite{PixelRec50K}
This dataset consists of nearly one million user-video interactions involving
around 50k users. Videos are represented through frame-level visual descriptors
and textual metadata. The dataset facilitates evaluation of fine-grained visual
signals in recommendation.

\textbf{GLAMI-1M.}
A multilingual fashion dataset spanning 13 languages and 191 categories,
containing one million product entries. Nodes are enriched with multilingual
descriptions and high-resolution images. Edges approximate semantic or
interaction-driven relations among items.

\textbf{MultiMET Series.}
The MultiMET collection includes graph-structured content from social platforms.
\textit{MultiMET facebook}~\cite{MultiMET} comprises approximately 8k posts, each containing text,
images, and temporal metadata, with edges reflecting user engagement.
\textit{MultiMET twitter}~\cite{MultiMET} offers about 7.5k multimodal tweets and their interaction
structure.
\textit{MultiMET ads}~\cite{MultiMET} includes around 6.8k advertising items, where textual and
visual content form the node attributes and edges approximate dissemination
patterns.

\textbf{Social Media, Vision-Language, and Medical Graphs}

\textbf{RedditS.}~\cite{RedditS}
This MAGB social-media dataset centers on Reddit posts and their relationships.
Each node encodes the textual content, visual media, and auxiliary statistics of
a post, while edges capture posting or co-occurrence patterns across threads.

\textbf{Flickr30k.}~\cite{flickr30k}
Comprising 31{,}783 images and 158{,}915 captions, this benchmark can be cast as
a region-phrase graph in which image regions and textual phrases are treated as
nodes, with edges modeling semantic alignment. It is widely used in retrieval
and grounding tasks.

\textbf{Multimodal Medical Imaging Dataset.}
More than 20k radiological images paired with clinical text form this dataset.
Nodes correspond to image regions and key clinical terms; edges encode semantic
associations between findings and textual descriptors. The dataset supports
studies on multimodal diagnosis and report generation.

\subsection{Baseline Description}
\label{sec:baseline_desc}
Given the uniqueness of MM-FGL, the baselines in our benchmark consist of four components:
(1) GNN backbones designed for centralized or local multimodal graph learning;
(2) standard FL originally proposed for graph-independent scenarios;
(3) heterogeneous FL methods that explicitly address system, model, or modality heterogeneity across clients;
and (4) graph foundation models that adopt a two-stage pretrain-then-finetune paradigm.

For component (1), considering that MM-GNN architectures vary substantially due to additional multimodal fusion/alignment designs, we include traditional GNN backbones to provide a stable and widely adopted structural reference point and representative MM-GNN backbones to cover explicit fusion/alignment mechanisms.The backbone GNN details implemented in our proposed MM-OpenFGL are listed below:

\textbf{GCN}~\cite{gcn} proposes a graph neural architecture based on a first-order approximation of spectral graph convolutions, learning node representations that jointly capture local topology and node attributes.

\textbf{GAT}~\cite{gat} introduces attention-based message passing to quantify the relative importance of neighbors during aggregation, enabling adaptive weighting over neighborhood information without relying on fixed, pre-defined graph priors.

\textbf{GraphSAGE}~\cite{graphsage} learns node embeddings by aggregating sampled neighbor features, providing an inductive framework that generalizes to unseen nodes/graphs by leveraging node attribute information.

\textbf{GIN}~\cite{gin} adopts a simple yet powerful message-passing architecture and is proven to be maximally expressive among common GNN variants, achieving representational power comparable to the Weisfeiler-Lehman graph isomorphism test.

\textbf{ChebNet}~\cite{chebnet} introduces a spectral graph convolution framework based on Chebyshev polynomial approximations of the graph Laplacian, where convolutional filters are parameterized as truncated Chebyshev expansions to enable strictly localized $K$-hop filtering while avoiding explicit eigen-decomposition, resulting in efficient and scalable graph convolutions with linear complexity in the number of edges.


\textbf{RevGAT}~\cite{revgat} extends GAT with reversible architectures to improve memory efficiency, allowing much deeper attention-based GNNs to be trained without storing layer-wise activations, while maintaining competitive representation capacity.


\textbf{MMGCN}~\cite{mmgcn} extends graph convolution to multimodal-attributed graphs by constructing modality-aware user-item bipartite graphs and performing message passing independently within each modality, followed by coordinated fusion through shared ID embeddings. This design enables the model to capture modality-specific structural signals and user preferences while preserving cross-modal consistency, making MMGCN a representative multimodal GNN backbone for centralized multimodal graph learning.

\textbf{MGAT}~\cite{mgat} extends graph attention networks to multimodal-attributed graphs by introducing modality-specific attention mechanisms that adaptively weight neighbors under different modalities, enabling effective fusion of heterogeneous modality information within attention-based message passing.

\textbf{GSMN}~\cite{gsmn} formulates multimodal data as structured graphs (e.g., objects/words as nodes with relations as edges) and performs cross-modal graph matching to learn fine-grained alignments between modalities. By jointly capturing both intra-modality structure and inter-modality correspondences through graph-based reasoning, GSMN provides a representative backbone for multimodal fusion and alignment on multimodal-attributed graphs.

\textbf{MGNet}~\cite{mgnet} is a multiplex graph neural network designed for multimodal graphs, which integrates tensor-based representation learning with graph convolution to jointly model intra-modality graph structures and cross-modality relationships. By leveraging tensor decomposition to construct a shared latent space and applying modality-specific GCN aggregators with learnable modality pooling, MGNet provides an effective backbone for multimodal fusion on multiplex graph data.

\textbf{MHGAT}~\cite{mhgat} is proposed for representation learning on graphs possessing both heterogeneous structural constraints and multimodal attributes. The model distinguishes itself by replacing manual meta-path selection with a two-stage aggregation scheme: neighbors are first clustered and pooled by relation type, then integrated into a shared feature space using an edge-level attention mechanism. Complementing this structural encoding, a modality-wise attention layer performs adaptive fusion of features such as visual and textual data. To ensure robust training and capture long-range dependencies, the architecture employs a dense connectivity pattern with residual shortcuts, effectively mitigating the over-smoothing phenomenon. Comparative evaluations across multiple datasets (e.g., IMDB, AMAZON) reveal that MHGAT achieves state-of-the-art performance in various downstream tasks, surpassing traditional heterogeneous graph neural networks.

\textbf{UniGraph2}~\cite{unigraph2} is a unified multimodal graph representation learning framework that encodes heterogeneous modalities into a shared graph space and learns modality-agnostic node representations via graph-based self-supervision. By abstracting diverse modalities into a unified graph structure, UniGraph2 serves as a general-purpose multimodal GNN backbone for multimodal-attributed graph learning.

\textbf{GraphMAE2}~\cite{graphmae2} is a self-supervised graph representation learning backbone that extends masked autoencoding to graphs by reconstructing masked node attributes and structures through graph neural networks. By leveraging reconstruction-based pretext tasks to learn transferable node representations, GraphMAE2 provides a strong general-purpose GNN backbone that can be adapted to multimodal and downstream graph learning scenarios.

For component (2), we include representative standard FL methods. These methods provide stable and widely adopted federated training protocols that can be directly plugged into graph and multimodal learning pipelines, enabling us to isolate the impact of multimodal and graph-specific designs from generic federated optimization effects. The standard FL baselines implemented in MM-OpenFGL are listed below:

\textbf{FedAvg}~\cite{fedavg} is a foundational FL algorithm that enables decentralized training across multiple clients while keeping data local. In each communication round, a central server broadcasts the global model, clients perform local optimization (e.g., SGD), and the server aggregates their updates via weighted averaging to refine the global model.

\textbf{FedProx}~\cite{fedprox} extends FedAvg by adding a proximal regularization term to the local objective, which stabilizes local training under statistical heterogeneity and systems variability (e.g., different amounts of local work), improving robustness on NonIID data.

\textbf{SCAFFOLD}~\cite{karimireddy2020scaffold} mitigates client drift by introducing control variates that correct local update directions. This variance-reduction style correction improves convergence behavior, particularly under heterogeneous data and partial client participation.

\textbf{MOON}~\cite{moon} is a model-contrastive FL framework that enhances local training by encouraging representation consistency between local and global models via contrastive objectives at the model/feature level.

\textbf{FedDC}~\cite{feddc} reduces local drift through lightweight parameter-level corrections: each client maintains an auxiliary variable to track the deviation between local and global parameters, promoting update consistency across rounds.

\textbf{FedExP}~\cite{jhunjhunwala2023fedexp} is a personalized federated learning method that explicitly models client-specific optimization preferences by learning adaptive aggregation weights. Instead of enforcing a single global update direction, FedExP dynamically balances global knowledge sharing and local personalization, improving convergence stability and performance under heterogeneous data distributions.

\textbf{FLASC}~\cite{flasc} improves communication efficiency in federated learning by sparsifying model updates only during transmission while preserving dense local training. This design enables significant communication reduction without introducing additional architectural assumptions, allowing FLASC to serve as a lightweight and effective optimization backbone in federated settings.

\textbf{GLOCALFAIR}~\cite{meerza2024glocalfair} is a fairness-aware federated learning method that jointly optimizes global performance and local fairness objectives. It introduces a global-local optimization scheme to mitigate bias amplification across clients while maintaining competitive overall accuracy.

\textbf{Calibre}~\cite{chen2024calibre} is a personalized federated learning framework that leverages self-supervised learning to train a global representation model while explicitly calibrating the trade-off between generality and personalization. By introducing client-specific prototype regularization and prototype-guided aggregation, Calibre improves both mean accuracy and fairness across heterogeneous clients under NonIID data distributions.

\textbf{FedGM}~\cite{fedgm} proposes a condensation-based federated graph learning paradigm that replaces model parameters or gradients with condensed graphs as the optimization carrier. Each client performs local subgraph condensation via gradient matching and uploads the condensed data to the server, which integrates them into a global condensed graph for training. This design effectively addresses subgraph heterogeneity while reducing communication overhead and privacy risks.

\textbf{FGC}~\cite{fgc} extends graph condensation to the federated setting by collaboratively synthesizing a compact condensed graph from distributed client subgraphs. It decouples the centralized gradient-matching process into client-side gradient computation and server-side aggregation, enabling the condensed graph to preserve global utility while respecting data locality. To address privacy risks introduced by releasing condensed graphs, FGC further incorporates information bottleneck principles to limit membership leakage, making it a representative baseline that bridges graph condensation and federated graph learning.

\textbf{FedSage+}~\cite{fedsage} extends FedSage to the subgraph federated learning setting by explicitly addressing missing cross-client neighbors. It jointly trains a GraphSAGE classifier with a local missing-neighbor generator that synthesizes potential cross-subgraph neighbors, enabling more complete neighborhood aggregation under federation and improving global generalization without sharing raw graph data.

\textbf{GCFL+}~\cite{gcfl} dynamically clusters clients based on GNN gradient patterns to mitigate structural and feature heterogeneity in Graph-FL settings. To handle fluctuating gradients, it further enhances GCFL with gradient-sequence clustering via dynamic time warping (DTW), improving clustering reliability and robustness.

\textbf{FedStar}~\cite{fedstar} shares structural embeddings across clients via an independent structure encoder. This design captures domain-invariant structural information while allowing personalized feature learning, mitigating feature misalignment and improving inter-graph federated learning performance.

\textbf{FedSPA}~\cite{Tan2025FedSPAG} addresses homophily heterogeneity in federated graph learning by explicitly modeling both homophily conflict and homophily bias across clients. It introduces Subgraph Feature Propagation Decoupling (SFPD) to separate homophilic and heterophilic message passing, enabling collaboration under unified homophily levels, and proposes Homophily Bias-Driven Aggregation (HBDA) to adaptively weight client contributions based on spectral and parameter-sensitivity cues, thereby improving global generalization.

\textbf{FedIIH}~\cite{fediih} addresses heterogeneity in federated graph learning by jointly modeling inter-client and intra-client heterogeneity. It infers subgraph distribution similarities via a hierarchical variational framework from a global perspective, while disentangling local subgraphs into multiple latent factors to enable factor-wise personalized federation, leading to robust collaboration across both homophilic and heterophilic graphs.

\textbf{FedSSP}~\cite{tan2024fedssp} proposes a personalized federated graph learning framework that addresses cross-domain structural heterogeneity from a spectral perspective. It shares generic spectral knowledge across clients to mitigate knowledge conflict induced by domain shifts, while retaining client-specific components to preserve personalization.

\textbf{FedHERO}~\cite{chen2025fedhero} addresses heterophily in federated graph learning by explicitly decoupling global structural patterns from client-specific graph characteristics. It learns a shared structure-oriented representation to capture common message-passing behaviors across clients, while allowing local models to preserve personalized neighborhood semantics, enabling effective collaboration under heterogeneous and heterophilic graph distributions.

\textbf{S2FGL}~\cite{tan2025s2fgl} tackles subgraph federated graph learning by jointly addressing spatial label-signal disruption and spectral client drift. It reinforces missing label semantics via prototype-based semantic sharing and aligns graph-frequency components across clients to improve robustness and generalization under heterogeneous subgraph distributions.

\textbf{FedLap}~\cite{fedlap} leverages global graph structure in subgraph federated learning via Laplacian smoothing, introducing a structural regularization term that implicitly enforces representation consistency among neighboring nodes without explicit message passing or feature sharing. By operating in the spectral domain, FedLap captures inter-subgraph dependencies while achieving strong privacy guarantees and low communication overhead.

\textbf{FedGLS}~\cite{fedgls} addresses subgraph heterogeneity in federated graph learning by aligning local representations through graph-level structure statistics. It introduces a global-local structure consistency objective that encourages clients to preserve shared structural patterns while maintaining personalized local embeddings, improving robustness under NonIID graph distributions.

\textbf{FedSheafHN}~\cite{Liang2024FedSheafHNPF} addresses heterogeneity in federated graph learning by modeling cross-client relationships through sheaf neural networks. It encodes local subgraphs as sheaf-based structures to capture consistent global dependencies while allowing client-specific variations, enabling effective collaboration under structurally diverse and NonIID graph distributions.

\textbf{FedDEP}~\cite{feddep} addresses data heterogeneity in federated graph learning by decoupling dependency modeling from parameter aggregation. It learns transferable dependency patterns across clients to guide local representation learning, enabling effective collaboration while preserving client-specific graph characteristics under NonIID settings.

\textbf{FCGL}~\cite{fcgl} addresses federated graph learning under client heterogeneity by performing collaborative graph representation learning with client-specific subgraph characteristics. It enables effective cross-client knowledge transfer through coordinated training while preserving local structural diversity, improving robustness and generalization in NonIID federated graph settings.

\textbf{FedAGHN}~\cite{Song2025FedAGHNPF} addresses heterogeneity in federated graph learning by adaptively modeling cross-client structural variations with heterogeneous neighborhood aggregation. It introduces a flexible aggregation mechanism that adjusts to diverse local graph patterns, enabling effective knowledge sharing while preserving client-specific structural characteristics under NonIID graph distributions.

For component (3), considering that real-world federated settings often involve  heterogeneity across clients, we include representative heterogeneous FL methods that explicitly address such discrepancies. The heterogeneous FL baselines implemented in our proposed MM-OpenFGL are listed below:

\textbf{FedProto}~\cite{tan2022fedproto} is an early prototype-based framework for heterogeneous federated learning. Instead of exchanging gradients, clients and the server share abstract class prototypes. FedProto aggregates local prototypes to form global prototypes and broadcasts them back to clients as regularizers, encouraging alignment between local representations and global prototype standards while reducing local classification errors.

\textbf{FedTGP}~\cite{zhang2024fedtgp} addresses model heterogeneity in federated learning by leveraging task-guided prototypes to facilitate knowledge transfer across heterogeneous client models. It aligns client-specific representations through prototype-level supervision rather than parameter sharing, enabling effective collaboration among clients with different architectures while preserving personalized model capacity.

\textbf{LG-FedAvg}~\cite{lg} addresses model heterogeneity in federated learning by decoupling global and local model updates. It maintains a shared global model to capture common knowledge across clients while allowing each client to preserve a personalized local model, enabling effective collaboration among heterogeneous client models without enforcing strict architectural consistency.

\textbf{FML}~\cite{fml} tackles model heterogeneity in federated learning by enabling flexible collaboration among clients with different model architectures. It facilitates knowledge transfer through representation-level alignment rather than direct parameter sharing, allowing each client to retain its own model structure while benefiting from shared federated information.

\textbf{FedKD}~\cite{fedKD} addresses model heterogeneity in federated learning by introducing knowledge distillation into the aggregation process. Instead of enforcing parameter-level consistency across heterogeneous client models, FedKD transfers knowledge via soft predictions or intermediate representations, enabling effective collaboration among clients with different architectures while preserving local model flexibility.

\textbf{FIARSE}~\cite{wu2024fiarse} addresses heterogeneity in federated learning by disentangling shared and client-specific representations through adaptive representation separation. It encourages clients to learn invariant features for global aggregation while preserving personalized components locally, thereby improving robustness under NonIID data and heterogeneous client distributions.

\textbf{ReeFL}~\cite{reefl} addresses system heterogeneity in federated learning by introducing a recurrent early-exit mechanism that enables clients with different resource budgets to train depth-adaptive sub-models. It employs a shared recurrent module to fuse representations from multiple exits into a single classifier and dynamically selects per-client teacher exits for knowledge distillation, yielding stable and efficient federation across heterogeneous clients.

\textbf{FedTSA}~\cite{fan2024fedtsa} tackles system and model heterogeneity in federated learning through a cluster-based two-stage aggregation framework. It first groups clients with similar resource capabilities to perform standard in-cluster aggregation, and then enables cross-cluster knowledge transfer among heterogeneous models via server-side deep mutual learning, using diffusion-model-generated data to avoid public datasets and support flexible, privacy-preserving aggregation.

\textbf{MH-pFLID}~\cite{mh-pflid} addresses model heterogeneity in federated learning by introducing an injection-distillation paradigm with a lightweight messenger model. Instead of relying on public datasets for knowledge distillation, MH-pFLID transfers knowledge across heterogeneous clients via injected and distilled representations, enabling efficient and privacy-preserving collaboration among models with different architectures.

\textbf{HAPFL}~\cite{hapfl} addresses device and model heterogeneity in federated learning via a dual-agent PPO framework that adaptively allocates heterogeneous model sizes and per-client training intensities. It further introduces a shared lightweight LiteModel with mutual knowledge distillation to enable effective global aggregation and mitigate stragglers.

\textbf{MH-pFedHNGD}~\cite{mh-pfedhngd} tackles model-heterogeneous personalized FL by using a server-side hypernetwork to generate client-specific parameters from learned embeddings, with a multi-head design that shares heads among clients with similar model sizes. It further adds a lightweight global model for an extra hypernetwork update and knowledge distillation to improve generalization without requiring external data or revealing client architectures. 

\textbf{PEPSY}~\cite{pepsy} addresses multimodal federated learning with both missing modalities and missing input features by learning client-side data-missing profiles that encode local missing patterns as embedding controls. These profiles are probabilistically aligned and aggregated on the server to reconfigure shared representations toward each client’s incomplete data view, enabling robust aggregation and stable performance under severe data incompleteness.

\textbf{FedMosaic}~\cite{fedmosaic} addresses both data and model heterogeneity in personalized federated learning by enabling relevance-guided aggregation that selectively shares knowledge among clients with related tasks, reducing interference. It further introduces PQ-LoRA, a dimension-invariant adaptation module that allows effective knowledge sharing across heterogeneous model architectures, improving personalization and generalization in realistic multimodal settings. 

\textbf{FedMVP}~\cite{fedmvp} addresses modality missing in multimodal federated learning by leveraging frozen pre-trained foundation models for cross-modal completion and representation knowledge transfer. It trains a lightweight joint encoder via multimodal contrastive objectives and performs CKA-based importance-aware aggregation on the server, achieving robust performance under severe modality incompleteness. 

\textbf{FedMAC}~\cite{nguyen2024fedmac} addresses cross-modal heterogeneity in multimodal federated learning by introducing a modality-aware collaboration framework that aligns heterogeneous modality representations via cross-modal contrastive learning. It further employs modality-specific aggregation with adaptive weighting to mitigate negative transfer across clients with inconsistent modality availability.

\textbf{FedMM}~\cite{peng2024fedmm} addresses modality heterogeneity in multimodal federated learning by federatedly training multiple single-modality feature extractors instead of a unified fusion model. It introduces prototype-guided representation alignment with a dynamic loss to enable effective collaboration across clients with partially overlapping modalities while preserving data privacy. 

\textbf{FediLoRA}~\cite{yang2025fedilora} addresses heterogeneous federated multimodal fine-tuning under missing modalities by enabling heterogeneous LoRA ranks across clients. It proposes a dimension-wise reweighted aggregation to prevent information dilution and a lightweight layer-wise LoRA editing strategy that repairs degraded local adapters using global parameters, improving both global and personalized performance. 

For component (4), we include representative GFMs to probe the potential of foundation-level backbones in multimodal federated graph learning. These models typically leverage large pretrained encoders and task-agnostic objectives, and are evaluated under a unified two-stage pipeline to assess their transferability and effectiveness in MM-FGL settings. The GFM backbones implemented in our proposed MM-OpenFGL include GFT, OFA, UniGraph, GQT, GraphCLIP, AnyGraph, GFSE, SwapGT, and RAGraph.

\textbf{GFT}~\cite{gft} is a graph foundation model that pretrains transformer-based architectures on large-scale graphs using self-supervised objectives to capture generic structural and semantic patterns. By decoupling pretraining from downstream tasks and adopting a pretrain-finetune pipeline, GFT provides transferable graph representations that can be adapted to diverse graph learning scenarios.

\textbf{OFA}~\cite{ofa} is a graph foundation model that unifies multiple graph learning tasks within a single pretrained transformer framework. By adopting a two-stage pretrain-finetune paradigm and learning task-agnostic graph representations through large-scale pretraining, OFA enables flexible adaptation to diverse downstream graph tasks and serves as a representative foundation-level backbone for graph learning.


\textbf{GQT}~\cite{gqt} is a graph foundation model that formulates graph learning as a sequence modeling problem and leverages large-scale pretraining with query-based representations. By following a two-stage pretrain-finetune paradigm, GQT learns generalizable structural and semantic patterns that can be transferred to downstream graph tasks.

\textbf{GraphCLIP}~\cite{graphclip} is a graph foundation model that extends CLIP contrastive pretraining to graph-structured data by jointly aligning graph representations with textual semantics. Following a two-stage pretrain-finetune paradigm, GraphCLIP learns transferable multimodal representations that can be adapted to downstream graph tasks.

\textbf{AnyGraph}~\cite{anygraph} is a graph foundation model designed to generalize across diverse graph structures and domains through large-scale pretraining. It adopts a two-stage pretrain-finetune paradigm to learn transferable representations that can be adapted to a wide range of downstream graph tasks, highlighting strong cross-graph and cross-domain generalization ability.

\textbf{GFSE}~\cite{gfse} is a graph foundation model that learns transferable graph representations through large-scale self-supervised pretraining. It follows a two-stage pretrain-finetune paradigm to capture general structural semantics, enabling effective adaptation to diverse downstream graph tasks.

\textbf{SwapGT}~\cite{swapgt} is a graph foundation model based on transformer architectures that leverages structure-aware pretraining objectives to enhance representation learning on graphs. By adopting a two-stage pretrain-finetune paradigm, SwapGT learns transferable graph representations that can be adapted to downstream graph tasks.

\textbf{RAGraph}~\cite{ragraph} is a graph foundation model that integrates retrieval-augmented generation into graph representation learning.
It adopts a pretrain-then-finetune paradigm, where external knowledge or retrieved graph contexts are incorporated during pretraining to enhance semantic grounding, enabling improved generalization and adaptability across downstream graph tasks without task-specific architectural redesign.


\subsection{Comparison with OpenFGL}
\label{appendix: Comparison with Related Benchmarks}

MM-OpenFGL represents a systematic evolution of the OpenFGL~\cite{Li2024OpenFGLAC} framework, inheriting its core design philosophy while broadening the application scope to multimodal federated graph learning. We maintain the modular API structure and standardized four-stage communication protocol introduced in OpenFGL to ensure a consistent environment for collaborative training. Our work advances the original single-modality benchmark by formalizing complex multimodal scenarios and introducing a tri-dimensional simulation strategy that addresses modality-specific challenges such as heterogeneity and missingness. Furthermore, MM-OpenFGL significantly expands the evaluation landscape with a dual-level task system (graph-level \& modality-level) and an extensive library of 57 state-of-the-art baselines, providing a more rigorous platform for assessing decentralized multimodal graph intelligence.

\subsection{Evaluation Protocols}
\label{appendix: metric}

To strictly benchmark diverse multimodal graph learning approaches, we standardize the experimental protocols across three core tasks: supervised node classification, supervised link prediction, and unsupervised node clustering. Across all experiments, we employ a frozen CLIP-ViT-Large-Patch14 as the universal feature encoder, fixing the feature dimension at 768. For optimization, node classification and clustering share a learning rate of $5 \times 10^{-3}$, a weight decay of $1 \times 10^{-5}$, and a batch size of 512. However, the training duration differs, with node classification requiring 100 epochs versus 500 epochs for clustering to ensure the convergence of self-supervised objectives. Conversely, the link prediction task is optimized with a learning rate of $1 \times 10^{-3}$ and a larger batch size of 2048 to accommodate extensive edge processing. 

To evaluate the model’s proficiency in cross-modal alignment within a unified latent space, we conduct modality retrieval tasks utilizing contrastive learning objectives integrated with a temperature scaling factor of $\tau = 0.07$. These retrieval models are optimized over 500 epochs with a learning rate of $1 \times 10^{-3}$ and a batch size of 256.
Furthermore, to bolster the generalizability of the learned representations and preemptively address overfitting, we implement an early stopping mechanism with a patience threshold ranging from 10 to 25 epochs. For modality generation, configurations are specifically calibrated to accommodate high-dimensional generative processes and the complex many-to-many dependencies inherent in the graph structure. 

In the G2Text task, the model is trained for 15 epochs using a learning rate of $1 \times 10^{-3}$, a weight decay of $1 \times 10^{-2}$, and a batch size of 8. This process leverages a self-attention with embeddings strategy to sample four multimodal neighbors while employing GNNs for structural position encoding. The decoder backbone is powered by the pre-trained Facebook OPT-125M, supporting both prefix tuning and LoRA with a rank of $r=64$ to ensure parameter efficiency during adaptation.

Similarly, the G2Image task is executed over 20 epochs with a learning rate of $1 \times 10^{-4}$ and a batch size of 16. Following the InstructG2I framework, we adopt semantic personalized PageRank-based neighbor sampling, selecting between 0 and 6 informative neighbors to provide necessary structural context. The image resolution is standardized at 256 to condition the Stable Diffusion v1.5 backbone through a Graph Classifier-Free Guidance mechanism. To maintain experimental rigor and minimize performance variance, core architectural parameters are standardized across all configurations. Unless otherwise specified, we employ CLIP-ViT-L/14 as the default feature encoder, with node embedding dimensions unified at 768 for all downstream tasks. All optimization processes are performed using the Adam or AdamW optimizers.

Based on this, the evaluation metrics are listed as follows:

\textbf{Accuracy (ACC)} measures the proportion of correctly classified nodes over the entire evaluation set. It computes the ratio between the number of correct predictions and the total number of samples. Let $N$ denote the number of nodes, $y_i$ and $\hat{y}_i$ the ground-truth and predicted labels of the $i$-th node, respectively, and $\mathbb{I}(\cdot)$ the indicator function. Accuracy provides an intuitive assessment of overall classification performance.
\begin{equation}
    \text{Acc} = \frac{1}{N} \sum_{i=1}^{N} \mathbb{I}(\hat{y}_i = y_i).
\end{equation}

\textbf{Precision} quantifies the reliability of positive predictions by measuring the fraction of correctly predicted positive samples among all samples predicted as positive. This metric highlights the model’s ability to avoid false positive errors and is particularly informative in scenarios where incorrect positive predictions are costly.
\begin{equation}
    \text{Precision} = \frac{\text{TP}}{\text{TP} + \text{FP}}.
\end{equation}

\textbf{Recall} evaluates the completeness of positive predictions by measuring the proportion of true positive samples that are successfully identified by the model. It reflects the ability to capture relevant instances and is essential for assessing performance on under-represented classes in node classification.
\begin{equation}
    \text{Recall} = \frac{\text{TP}}{\text{TP} + \text{FN}}.
\end{equation}

\textbf{F1-Score} is defined as the harmonic mean of Precision and Recall, offering a unified measure that balances false positives and false negatives. This metric is especially suitable for graph datasets with imbalanced label distributions, providing a more robust evaluation than Accuracy alone.
\begin{equation}
    \text{F1} = 2 \cdot \frac{\text{Precision} \cdot \text{Recall}}{\text{Precision} + \text{Recall}}.
\end{equation}

\textbf{Average Precision (AP)} summarizes the precision--recall curve into a single scalar by averaging precision values across different recall levels. It emphasizes ranking quality under class imbalance and is widely adopted for evaluating modality matching tasks where positive and negative pairs are unevenly distributed.
\begin{equation}
    \text{AP} = \sum_n (R_n - R_{n-1}) P_n,
\end{equation}
where $P_n$ and $R_n$ denote the precision and recall at the $n$-th threshold, respectively.

\textbf{Area Under the ROC Curve (AUC-ROC)} measures the probability that a randomly chosen positive pair is assigned a higher matching score than a randomly chosen negative pair. By integrating the true positive rate against the false positive rate over all possible thresholds, AUC-ROC provides a threshold-independent evaluation of matching robustness.
\begin{equation}
    \text{AUC} = \int_0^1 \text{TPR}(t) \, d\,\text{FPR}(t).
\end{equation}

\textbf{Recall@K} evaluates retrieval effectiveness by measuring the fraction of queries for which the correct target appears within the top-$K$ ranked results. This metric reflects the system’s practical utility under a fixed retrieval budget.
\begin{equation}
    \text{Recall@}K = \frac{1}{|Q|} \sum_{i=1}^{|Q|} \mathbb{I}(\text{rank}_i \le K),
\end{equation}
where $Q$ denotes the set of queries and $\text{rank}_i$ the rank of the ground-truth item for the $i$-th query.

\textbf{Mean Reciprocal Rank (MRR)} assesses the overall quality of ranked retrieval results by considering the inverse rank of the first correct item. It rewards models that consistently place relevant targets at higher positions in the ranking list and complements Recall@K by accounting for exact rank placement.
\begin{equation}
    \text{MRR} = \frac{1}{|Q|} \sum_{i=1}^{|Q|} \frac{1}{\text{rank}_i}.
\end{equation}

\textbf{BLEU} evaluates the quality of generated text by measuring the modified $n$-gram precision between generated outputs and reference texts, together with a brevity penalty to discourage overly short generations. BLEU reflects lexical accuracy and surface-level fluency.
\begin{equation}
    \text{BLEU} = \text{BP} \cdot \exp\left(\sum_{n=1}^{N} w_n \log p_n\right).
\end{equation}

\textbf{ROUGE-L} measures generation quality based on the longest common subsequence between generated and reference texts. Unlike fixed $n$-gram metrics, ROUGE-L captures sentence-level structure and emphasizes recall, ensuring that the generated content sufficiently covers key information.
\begin{equation}
    \text{ROUGE-L} = \frac{(1 + \beta^2) R_{lcs} P_{lcs}}{R_{lcs} + \beta^2 P_{lcs}}.
\end{equation}

\textbf{CIDEr} evaluates the consensus between a generated description and multiple reference texts using TF-IDF weighted $n$-gram similarity. By down-weighting common phrases and emphasizing informative terms, CIDEr better aligns with human judgment.
\begin{equation}
    \text{CIDEr}_n(c, r) = \frac{1}{M} \sum_{i=1}^{M} 
    \frac{g^n(c) \cdot g^n(r_i)}{\|g^n(c)\| \|g^n(r_i)\|}.
\end{equation}

\subsection{Theoretical Complexity Analysis}
\label{appendix: complexity}

    In this section, we provide a detailed analysis of the theoretical time and space complexities.
    The symbol descriptions are as follows: $n$, $m$, and $c$ denote the number of nodes, edges, and classes, respectively. 
    $K$ corresponds to the number of times we aggregate features. $h$ represents the dimension of the hidden layer. 
    $f$ and $F$ represent the feature dimension of the default backbone and the model additionally constructed by the method. 
    $q$ represents the number of eigenvalues of the feature map. 
    $E$ represents the number of times to construct a sparse matrix.
    Based on this, our analysis is as follows:
    
    \textbf{Standard GNN Backbone Complexity}.
    The majority of the evaluated methods utilize a standard Graph Neural Network (e.g., GCN, GAT) as the local backbone. 
    The computational cost for a single client training round is determined by two primary operations: sparse neighborhood aggregation and dense feature transformation. 
    For a model with $K$ layers, the aggregation step involves message passing along $m$ edges with feature dimension $f$, resulting in a time complexity of $O(K \cdot m \cdot f)$. The feature transformation step involves linear mapping of $n$ nodes with weight matrices of size $f \times f$, contributing $O(n \cdot f^2)$. 
    Combining these yields the standard time complexity term $O(Kmf + nf^2)$. 
    Similarly, the space complexity is dominated by the need to store node embeddings for $n$ nodes across layers for gradient computation, resulting in $O(nf^2)$. 
    This baseline complexity applies directly to methods like FedMVP, FedSSP, and standard baselines, where no additional heavy structural or model modules are introduced.
    
    \textbf{Model Heterogeneous Methods}.
    Methods such as FML, MH-pFLID, and PEPSY address heterogeneity by maintaining dual architectures or auxiliary embedding profiles. 
    For FML and MH-pFLID, the learning process involves a primary local model with feature dimension $f$ and a secondary global or messenger model with dimension $F$. 
    While the training time is asymptotically dominated by the graph operations of the primary backbone, yielding $O(Kmf + nf^2)$, the memory requirement is additive.
    The system must store parameters and gradients for both the local architecture and the auxiliary model, leading to the space complexity $O(N(f^2 + F^2))$ reported in the table.~\ref{tab: complexity}.
    In the specific case of PEPSY, the method learns client-specific missing profiles. 
    These profiles act as learnable embeddings that must be aligned with both the input space and the projected latent space, necessitating storage proportional to both $f^2$ and $F^2$ for all $N$ nodes, thus matching the $O(N(f^2 + F^2))$ memory complexity.
    
    \textbf{Structural Learning Methods}.
    Methods such as FedSPA, FedIIH, and FedLap introduce mechanisms to handle Non-IID graph structures, which often require dense matrix operations that exceed the sparsity of the original graph. FedSPA involves updating a structural matrix to capture global homophily patterns. 
    The term $En^2$ in its time complexity $O(Kmf + nf^2 + En^2)$ represents the cost of these structural updates, where $E$ denotes the number of update iterations and $n^2$ reflects the dense nature of the global structural approximation. 
    Consequently, the space complexity includes an $O(n^2)$ term to store this dense structure. 
    Similarly, FedIIH models inter-client heterogeneity via a hierarchical variational framework that infers latent distribution factors. This inference process involves pairwise interactions between nodes to reconstruct structural dependencies, adding an $O(n^2)$ overhead to both time and space complexity. 
    FedLap utilizes Laplacian regularization; while its time complexity remains dominated by the backbone due to efficient implementation, the theoretical space complexity includes $O(n^2)$ to account for the storage of the dense affinity or Laplacian matrix required for the regularization term.
    
    \textbf{Complexity of Spectral and Prototype Methods}.
    S2FGL addresses heterogeneity through spatial-spectral consistency, which requires eigen-decomposition or spectral projections. 
    The time complexity $O(Kmf + nf^2 + qn^2)$ includes the standard GNN terms plus $qn^2$, which corresponds to the projection of $n$-dimensional graph signals onto $q$ selected eigenvectors. The space complexity $O(Nf^2 + n^2 + qn)$ accounts for the node features ($Nf^2$), the dense basis or correlation matrix ($n^2$), and the storage of the top-$q$ eigenvectors ($qn$).
    On the efficiency side, FedMAC employs a modality-aware collaboration scheme using prototypes. 
    Its space complexity is $O(Nf^2 + hc)$, where the term $hc$ represents the storage required for class-wise prototypes (where $h$ is the hidden dimension and $c$ is the number of classes). 
    This term is significantly smaller than the node-level storage, highlighting the method's memory efficiency compared to structural learning approaches.

\end{document}